%% file: main.tex
\documentclass[journal]{IEEEtran}
\ifCLASSINFOpdf
\else
\fi

\usepackage[utf8]{inputenc}
\usepackage{algorithm}
\usepackage{algpseudocode}
\usepackage{makecell}
\usepackage{amssymb}
\usepackage{fullpage}
\usepackage{multirow} 
\usepackage{amsmath}
\usepackage{autobreak}
\usepackage{graphicx}
\usepackage{subfigure}
\usepackage{caption}
\title{A Critical Review of Inductive Logic Programming Techniques for Explainable AI}
\author{Zheng~Zhang,
        Liangliang~Xu,       Levent~Yilmaz,~\IEEEmembership{Member,~IEEE,} and 
        Bo~Liu,~\IEEEmembership{Senior Member,~IEEE}
\thanks{Z. Zhang, L. Xu, L. Yilmaz, and B. Liu are with the Department of Computer Science and Software Engineering, Auburn University, Auburn, AL, 36839, USA.} 
\thanks{Corresponding author: Bo Liu $<$boliu@auburn.edu$>$.}}

\date{}

\usepackage{color, soul}
\soulregister\cite7
\soulregister\ref7
\soulregister\$7
\setulcolor{blue}
\setstcolor{red}

\DeclareUnicodeCharacter{2192}{-}

\begin{document}
	
	\maketitle   
	\begin{abstract}
        Despite recent advances in modern machine learning algorithms, the opaqueness of their underlying mechanisms continues to be an obstacle in adoption. To instill confidence and trust in artificial intelligence systems, Explainable Artificial Intelligence has emerged as a response to improving modern machine learning algorithms' explainability. Inductive Logic Programming (ILP), a subfield of symbolic artificial intelligence, plays a promising role in generating interpretable explanations because of its intuitive logic-driven framework. ILP effectively leverages abductive reasoning to generate explainable first-order clausal theories from examples and background knowledge. However, several challenges in developing methods inspired by ILP need toç be addressed for their successful application in practice. For example, existing ILP systems often have a vast solution space, and the induced solutions are very sensitive to noises and disturbances. This survey paper summarizes the recent advances in ILP and a discussion of statistical relational learning and neural-symbolic algorithms, which offer synergistic views to ILP. Following a critical review of the recent advances, we delineate observed challenges and highlight potential avenues of further ILP-motivated research toward developing self-explanatory artificial intelligence systems.
		~\\
		
		\noindent \textbf{keywords:} 
		differentiable inductive logic programming, neuro-symbolic AI, inductive logic programming, meta-interpretive learning, machine learning, probabilistic inductive logic programming, statistical relational learning, XAI.
		
	\end{abstract}    
	
    \input{sec0102}

    \input{sec030405}
    \input{sec060708}

	\bibliographystyle{ieeetr}
	\bibliography{ILP_survey_ref}
	
    \newpage
    \input{biography}
\end{document}

%% file: sec0102.tex
    \section{Introduction}
    \label{sec:intro}
    Explainability has become an important research area to overcome the challenges in addressing the complexity and understandability of Artificial Intelligence (AI) systems \cite{adadi2018peeking}. 
    XAI is an emerging field \cite{arrieta2020explainable} in machine learning that refers to methods and techniques that enable experts to understand the decisions made by AI algorithms. The US Defense Advanced Research Project Agency describes three AI explainability criteria: prediction accuracy, decision understanding or trust, and traceability \cite{gunning2019darpa}. Prediction accuracy refers to explaining how specific conclusions are derived, while decision understanding involves fostering trust in the underlying mechanisms and processes. Traceability involves inspection of the actions and their cause-effect relations to improve causal and evidential reasoning and to immerse into the decision loops to influence tasks as needed. According to these criteria, an AI system should perform a specific task or recommend decisions and produce an explainable characterization of why it renders specific decisions along with the supporting rationale. 
    
\begin{figure*}[h]
		\centering
		\centerline{\includegraphics[width=30pc,height=10pc]{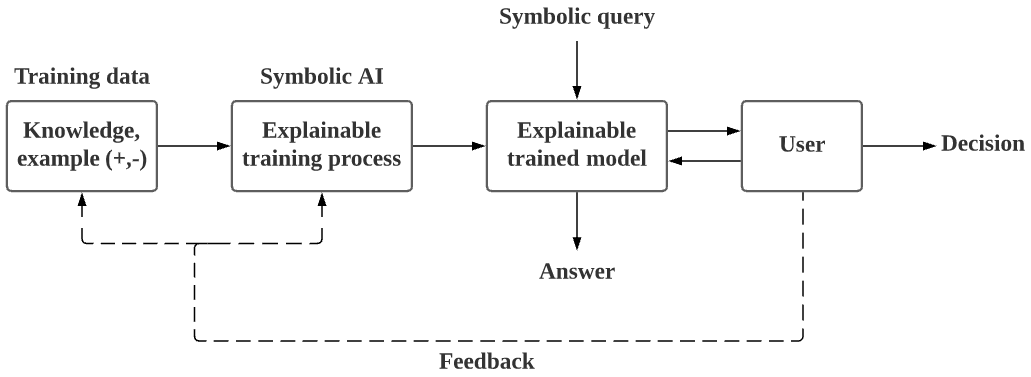}}
		\caption{A flowchart of the Symbolic AI pipeline}
		\label{fig:symbolic_AI_diagram}
	\end{figure*}

    Recently, with the success of precise but largely inscrutable deep learning models, explainability received significant attention \cite{furnkranz2019on}. The opaqueness of deep learning algorithms motivates AI researchers to open the black box for bringing transparency and avoiding reliance only on model accuracy \cite{adadi2018peeking}. Explanations are essential for the explainability of a machine-learned model. Symbolic Artificial Intelligence (Symbolic AI) is the overarching framework for AI methods that are based on high-level ``symbolic" (human-readable) representations of problems, plans, and solutions. Symbolic AI requires a modest amount of training and can achieve reasonable performance with limited data. Symbolic methods also grant the benefits of explainability and generalization at the concept level through inferential and inductive reasoning while offering connections to established planning algorithms~\cite{calegari2020logic}. Fig. \ref{fig:symbolic_AI_diagram} shows a flowchart of the Symbolic AI pipeline: by learning from symbolic background knowledge and examples (positive and negative), the symbolic model can elaborate and revise logic frames in an explainable manner while enabling the provision of valid and provable answers for symbolic queries. The user could interact with the trained system as follows: the system can take action for the current task and provide an explanation to the user that justifies its action; the user can then make a decision based on the explanation and improve the knowledge base as well as the training process.

    As Symbolic AI achieves unprecedented impact in problem-solving via search, planning, and decision-making, one of the hallmarks of Symbolic AI is Inductive Logic Programming (ILP) \cite{cropper2020inductive}. Besides explainability, ILP systems are also data-efficient, while most machine learning models require large quantities of data to reach acceptable accuracy during learning. ILP, as a knowledge-based strategy, can provide a symbolic, goal-driven, and causal interpretation of data. 
    
    However, there are also some drawbacks to these symbolic approaches: they are inherently brittle and do not deal well with noise, it is not easy to express complex non-linear decision surfaces in logic \cite{schmid2018inductive}, and the guiding method for hypothesis search is questionable \cite{cropper2020learning}, etc. To improve the efficacy of ILP methods and mitigate the challenges, several types of ILPs have been developed: meta-interpretive learning (MIL) framework \cite{muggleton2014meta} has made progress in predicate invention and recursive generalizations using abduction in terms of a meta-interpreter; probabilistic inductive logic programming (PILP) that introduced by De Raedt and Kersting \cite{de2008probabilistic} is more powerful than ILP and, in turn, traditional attribute-value approaches due to its ability to deal explicitly with the uncertainty; differentiable inductive logic programming ($\partial$ILP) \cite{evans2018learning} augments ILP with neural networks, making the system robust to noise and error in the training data that ILP cannot cope with alone. 
	
    The rest of the paper is organized based on the flowchart of the symbolic AI pipeline. In Section~\ref{sec:background}, related works in the extant literature regarding ILP are reviewed. Section~\ref{sec:variants-of-ilp} discusses three types of ILP, including PILP, MIL, and $\partial$ILP. Section~\ref{sec:data-driven-predicate-extraction} introduces symbolic-based frameworks, including statistical relational learning (SRL) and neural-symbolic AI, and their relation. In section~\ref{sec:user-centered_explainable_ilp}, We present a user-centered explanation and a measurement of trust appropriate for ILP. The experimental evaluations of ILP systems are depicted in Section~\ref{sec:experiments}. Section~\ref{sec:challenges} delineates the challenges and highlights future research directions for academics and domain experts, and section~\ref{sec:conclusion} concludes the paper.

	\section{Background}  
	\label{sec:background}
	Inductive logic programming (ILP), a classical rule-based system, is a subfield of symbolic artificial intelligence that uses logic programming as a uniform representation, for example, background knowledge and hypotheses. An ILP system will derive a hypothesized logic program, which entails all the positive and none of the negative examples given an encoding of the general background knowledge and a set of examples representing a logical database of facts.
	
	ILP was first introduced in \cite{muggleton1991inductive}. The successes of ILP have been in the area of inductive construction of expert systems: MYCIN \cite{shortliffe1975model} and XCON \cite{fox1986role} are built using hand-coding of rules; GASOIL \cite{slocombe1986engineering}, and BMT \cite{hayes1990news} are built using software derived from Quinlan's inductive decision tree building algorithm ID3 \cite{quinlan1979discovering}. Along with the successes of this technology, the following limitations have become apparent: 1. Propositional level systems cannot be used in areas requiring essentially relational knowledge representations. 2. Inability to make use of background knowledge when learning. 3. The systems construct hypotheses within the limits of a fixed vocabulary of propositional attributes \cite{muggleton1991inductive}.
	
	Later, Plotkin \cite{plotkin1972automatic} and Shapiro \cite{shapiro1982algorithmic} introduce computer-based inductive systems within the framework to solve these problems of full first-order logic. Plotkin's significant contributions are 1. the introduction of relative subsumption, a relationship of generality between clauses, and 2. the inductive mechanism of relative least general generalization (RLGG). Shapiro~\cite{shapiro1982algorithmic} investigates an approach to Horn clause induction in which the search for hypotheses is from general to specific, rather than Plotkin's specific to the general approach. These all become the foundation of ILP. Later, Sammut and Banerji \cite{sammut1986learning} describe a system called MARVIN, which generalizes a single example at a time regarding a set of background clauses. Quinlan \cite{quinlan1989learning} has described a highly efficient program called Foil, which relies on a general to specific heuristic search, which is guided by an information criterion related to entropy. Muggleton and Feng \cite{muggleton1990efficient} apply a ``determinate" restriction to hypotheses, and their learning program, Golem, has been demonstrated \cite{muggleton1990efficient} to have a level of efficiency similar to Quinlan's Foil but without the accompanying loss of scope. In the late 80s, right before ILP appears, Muggleton and Buntine \cite{muggleton1988machine} present Inverse Resolution (IR) and implement it in Cigol \cite{muggleton1988machine}. They also introduce a mechanism for automatically inventing and generalizing first-order Horn clause predicates. 
	
	\begin{figure*}[tp]
		\centering
		\captionsetup{justification=centering}
		\includegraphics[width=\textwidth,height=4in]{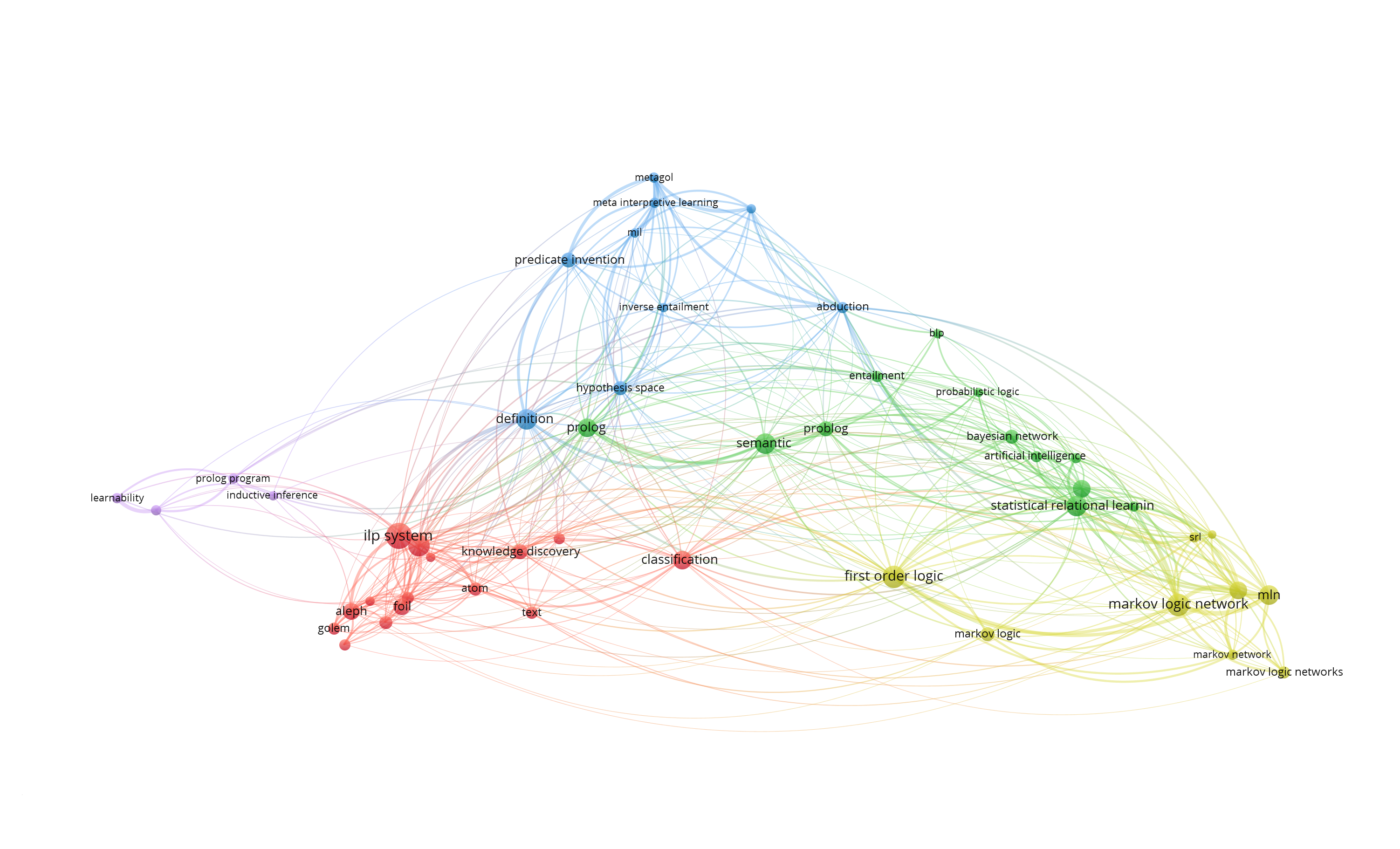}
		\caption{ILP as Transdisciplinary Concept}
		\label{fig:ILP-system-related}
	\end{figure*}
	
	ILP produces a widely used technology with a firm theoretical foundation based on principles from both Logic and Statistics. On the side of statistical justification of hypotheses, Muggleton \cite{muggleton2012ilp} discusses the possible relationship between Algorithmic Complexity theory and Probably-Approximately-Correct (PAC) Learning; in terms of logic, Muggleton provides a unifying framework for IR and RLGG by rederiving RLGG in terms of IR. Great progress has been made since the birth of ILP. Muggleton introduces more comprehensive approaches to first-order logic (FOL) inductive theory: inverting implication and inverse entailment. Kietz \cite{kietz1993some} proves that ILP generally is not PAC-learnable. Nienhuys-Cheng and Wolf \cite{nienhuys1997foundations} show the theoretical basis of ILP. Besides theoretical works, some ILP algorithms and implementations have also been developed: Progol is Stephen Muggleton's implementation of inductive logic programming that combines ``Inverse Entailment" with the ``general-to-specific search" through a refinement graph; Golem \cite{muggleton1990efficient}, developed by Stephen Muggleton and Feng, is based on RLGG; Lavrac et al. present an ILP system called LINUS based on propositional logic; CLINT uses queries to eliminate irrelevant literals and raises the generality of hypotheses by proposing more complex hypothesis clauses. Along with the era of machine learning and data-driven artificial intelligence, ILP naturally becomes an important part of the ML model because the learned hypotheses of ILP are represented in a symbolic form. They are inspectable by humans and therefore provide transparency and comprehensibility of the machine-learned classifiers. 

	We conclude our introduction with fig. \ref{fig:ILP-system-related} that depicts the overlay visualization of keywords related to ILP based on the analysis of over 300 articles. The term ILP system is associated with articles across multiple clusters. The articles that use ILP to develop systems have strong connections with the terms first-order logic, Aleph \cite{srinivasanaleph}, Foil \cite{quinlan1990learning}, and Prolog \cite{bratko2001prolog}. The high-order logic abduction method connects the term ILP system with the terms meta-interpretive learning \cite{muggleton2014meta}, Metagol \cite{metagol}, and predicate invention. Common terms that combine probability with the term ILP system are statistical relational learning, Markov logic network, and bayesian network. Keywords, including learnability, inductive inference, and classification, also have close relations to the term ILP system.
	
	\subsection{Concepts of ILP}
	We use the following example to introduce the basic concepts of ILP system:\\
	
	$$parent(i,a) \quad parent(a,b)$$ 
	$$grandparent(X,Y) \leftarrow parent(X,Z), parent(Z,Y)$$
	
	ILP is defined using first-order logic \cite{smullyan1995first}. In first-order logic, a formula, $parent(i,a)$ or $parent(a,b)$, that contains no logical connectives is called the atom. An atom or its negation, i.e. $\neg parent(i,a)$ and $\neg parent(a,b)$, is called literal. A definite clause is a many-way OR (disjunction) of literals (formula 3). $a,b,i$ in the example are constants, and X, Y, Z are variables, and all constants and variables are terms. A term that does not contain any free variables is called a ground term ($parent(i, a)$). A Boolean-valued function P: $X \rightarrow \{true, false\}$, is called the predicate ($parent$ and $grandparent$) on X. $grandparent/2$, $parent/2$ denote predicates with their arity, i.e., number of arguments. A function can be any value unlike a predicate, and it will never appear except as arguments to predicates. First-order logic is a structure of logic consisting of constants, variables, predicates, functions, and sentences. It uses quantified variables over non-logical objects and allows the use of sentences that contain variables.
	
	\subsection{Semantics of ILP}
	There are two different semantics for ILP: standard and non-monotonic semantics. For the normal semantic, given background (prior) knowledge $B$ and example $E$ ($E = E^+ \wedge E^-$ consists of positive example $E^+$ and negative example $E^-$), find a hypothesis $H$ such that the following conditions hold. 
	
	$$\textbf{Prior Satisfiability.}: B \wedge E^- \nvDash \Box$$
	$$\textbf{Posterior Satisfiability.}: B \wedge H \wedge E^- \nvDash \Box$$
	$$\textbf{Prior Necessity.}: B \nvDash E^+$$
	$$\textbf{Posterior Sufficiency.}: B \wedge H \vDash E^+$$
	
	A general setting is used for the normal semantics. In most ILP systems, the definite setting will be used as a simple version of the normal setting, as the background theory and hypotheses are restricted to being definite. The example setting is a special case of definite semantics, where the example is restricted to true and false ground facts. Notice that the example setting is equivalent to the normal semantics, where $B$ and $H$ are definite clauses and $E$ is a set of ground unit clauses. The example setting is the main setting of ILP. It is employed by the large majority of ILP systems. Table \ref{grandparent} shows the $grandparent$ dataset for the ILP system. The task is to learn the $grandparent$ relation from various facts involving the father-of and mother-of relations:
	
	\begin{table}[htbp]
		\centering
		\caption{$grandparent$ dataset}
		\begin{tabular}{lllll}
			\cline{1-3}
			\multicolumn{2}{|c|}{\textbf{Examples}} & \multicolumn{1}{c|}{\multirow{2}{*}{\textbf{Background Knowledge}}}                                                                                                                   &  &  \\ \cline{1-2}
			\multicolumn{1}{|c|}{$E^{+}$} & \multicolumn{1}{c|}{$E^{-}$}                                                                                                                                                                                                          & \multicolumn{1}{c|}{}                                                                                                                                                                  &  &  \\ \cline{1-3}
			\multicolumn{1}{|c|}{\begin{tabular}[c]{@{}l@{}}grandparent(i,b)    \\ grandparent(i,c)    \\ grandparent(a,d)    \\ grandparent(a,e)        \\ grandparent(a,f)        \\ grandparent(a,g)    \\ grandparent(c,h)\end{tabular}} & \multicolumn{1}{c|}{\begin{tabular}[c]{@{}l@{}}grandparent(a,b)        \\ grandparent(b,c)        \\ grandparent(c,d)        \\ grandparent(d,e)        \\ grandparent(e,f)        \\ grandparent(f,g)        \\ grandparent(g,h)        \\ grandparent(h,i)\end{tabular}} & \multicolumn{1}{c|}{\begin{tabular}[c]{@{}l@{}}father(a,b)        \\ father(a,c)        \\ father(b,d)        \\ father(b,e)\\ mother(i,a)        \\ mother(c,f)        \\ mother(c,g)        \\ mother(f,h)\end{tabular}} &  &  \\ \cline{1-3}
		\end{tabular}
		\label{grandparent}
	\end{table} 
	
	In the non-monotonic setting of ILP, the background theory is a set of definite clauses, the evidence is empty, and the hypotheses are sets of general clauses expressible using the same alphabet as the background theory. The reason that the evidence is empty is that the positive evidence is considered part of the background theory, and the negative evidence is derived implicitly by making a kind of closed world assumption (realized by taking the least Herbrand model \cite{ebbinghaus2013mathematical}). The non-monotonic semantics realizes induction by deduction. The induction principle of the non-monotonic setting states that the hypothesis $H$, which is, in a sense, deduced from the set of observed examples $E$ and the background theory B (using a kind of closed world and closed domain assumption), holds for all possible sets of examples. This produces generalizations beyond the observations. As a consequence, properties derived in the non-monotonic setting are more conservative than those derived in the normal setting.  
	
	\subsection{Searching method}
	An enumeration algorithm will be used to solve the ILP problem. Generalization and specialization form the basis for pruning the search space. Generalization corresponds to induction, and specialization to deduction, implying that induction is viewed here as the inverse of deduction. A generic ILP system can now be defined: 
	
	\begin{algorithm}
		\caption{searching algorithm \cite{muggleton1994inductive}}
		\begin{algorithmic}
			\State $QH:= Initialize$
			\Repeat 
			\State Delete $H$ from $QH$
			\State Choose the inference rules $r_1$, .... $r_k \in R$ to be applied to $H$
			\State Apply the rules $r_1$, .... $r_k$ to $H$ to yield $H_1, .... H_n$
			\State Add $H_1, .... H_n$ to $QH$
			\State Prune $QH$
			\Until{stop-criterion$(QH)$ satisfied}
		\end{algorithmic}
		\label{alg:searching-alg}
	\end{algorithm}
	
	$QH$ denotes a queue of candidate hypotheses. The algorithm \ref{alg:searching-alg} works as follows: It keeps track of $QH$, then repeatedly deletes a hypothesis $H$ from the queue and expands the hypotheses using inference rules. Then, the expanded hypotheses are added to the queue of hypotheses $QH$, which may be pruned to discard the unpromising hypothesis from further consideration. This process continues until the stop-criterion is satisfied.
	
	There are two kinds of search methods for ILP systems: ``specific-to-general" systems, a.k.a. bottom-up systems, start from the examples and background knowledge and repeatedly generalize their hypothesis by applying inductive inference rules. During the search, they take care that the hypothesis remains satisfiable (i.e., does not imply negative examples). ``general-to-specific” systems, a.k.a. top-down systems, start with the most general hypothesis (i.e., the inconsistent clause $\Box$) and repeatedly specializes the hypothesis by applying deductive inference rules to remove inconsistencies with the negative examples. During the search, care is taken that the hypotheses remain sufficient with regard to the positive evidence. Table \ref{search} shows some related systems for both two types:
	
	\begin{table}[htbp]
		\centering
		\caption{ILP systems with different searching methods}
		\begin{tabular}{|p{0.4\columnwidth}|p{0.4\columnwidth}|}
			\hline
			\textbf{specific-to-general} & \textbf{general-to-specific}\\
			\hline
			CLINT \cite{de1991clint}, Golem \cite{muggleton1990efficient}, Cigol \cite{muggleton1988machine} & MIS \cite{shapiro1982algorithmic}, Foil, Progol \cite{muggleton1995inverse}, LINUS \cite{lavravc1991learning}, Aleph\\
			\hline
		\end{tabular}
		\label{search}
	\end{table}    
	
	\subsection{Inductive inference rules}
	
	Induction can be considered as the inverse of deduction. Given the formulas $B \wedge H \models E^+$, deriving $E^+$ from $B \wedge H$ is deduction, and deriving $H$ from $B \wedge E^+$ is induction. Therefore, inductive inference rules can be obtained by inverting deductive ones. Table \ref{de_n_in} summarizes commonly used rules for both deduction and induction.
	
	\begin{table}[ht]
		\centering
		\caption{Deduction and Induction operations}
		\begin{tabular}{|c|c|}
			\hline
			\textbf{Deduction} & \textbf{Induction} \\ \hline
			\multicolumn{1}{|c|}{Most General Unification} & \multicolumn{1}{c|}{Least General Generalization} \\
			\multicolumn{1}{|c|}{Resolution}               & \multicolumn{1}{c|}{Inverse Resolution}           \\ 
			\multicolumn{1}{|c|}{Implication}              & \multicolumn{1}{c|}{Inverse Implication}          \\
			\multicolumn{1}{|c|}{Entailment}               & \multicolumn{1}{c|}{Inverse Entailment}           \\ \hline
		\end{tabular}
		\label{de_n_in}
	\end{table}
	
	Since this “inverting deduction” paradigm can be studied under various assumptions, corresponding to different assumptions about the deductive rule for $\vDash$ and the format of background theory $B$ and evidence $E^+$, different models of inductive inference are obtained. Four frameworks of inference rules, $\theta - subsumption$, Inverse Resolution, Inverse Implication, and Inverse Entailment, will be described in this section. 
	
	\textbf{$\theta$  - subsumption}. The $\theta$  - subsumption inductive inference rule is: 
	
	$$\theta  - subsumption: \frac{c^2}{c^1}\quad where \quad c^1 \theta \subseteq c^2$$
	
	For example, $grandparent(X, Z) \leftarrow parent(X, Y),parent(Y, Z)$ $\theta$-subsumes $grandparent$ $(i, b)\leftarrow parent(i, $ $ a), parent(a, b)$ with $\theta = \{X = i, Y = a, Z = b\}$. In the simplest $\theta-subsumption$, the background knowledge is supposed to be empty, and the deductive inference rule corresponds to $\theta-subsumption$ among single clauses. One extension of $\theta-subsumption$ that takes into account background knowledge is called relative subsumption. Like $\theta - subsumption$, it is straightforward to define relatively reduced clauses using a straightforward definition of relative clause equivalence. Relative subsumption forms a lattice over relatively reduced clauses. 
	
	\textbf{Inverse Resolution}. Inductive inference rules can be viewed as the inverse of deductive rules of inference. Since the deductive rule of resolution is complete for the deduction, an inverse of resolution should be complete for induction. Inverse Resolution takes into account background knowledge and aims at inverting the resolution principle. Four main rules of Inverse Resolution are widely used:
	
	$$\textbf{Absorption}: \qquad\frac{q \leftarrow A \qquad p \leftarrow A, B}{q \leftarrow A \qquad p \leftarrow q, B}$$
	
	$$\textbf{Identification}: \qquad\frac{p \leftarrow A, B \qquad p \leftarrow A, q}{q \leftarrow B \qquad p \leftarrow A, q}$$
	
	$$\textbf{Intra-Construction}: \qquad\frac{p \leftarrow A, B \qquad \qquad p \leftarrow A, C}{q \leftarrow B \quad p \leftarrow A, q \quad q \leftarrow C}$$
	
	$$\textbf{Inter-Construction}: \qquad\frac{p \leftarrow A, B \qquad \qquad q \leftarrow A, C}{p \leftarrow r, B \quad r \leftarrow A \quad q \leftarrow r, C}$$
	
	In these rules, lower-case letters are atoms, and upper-case letters are conjunctions of atoms. Both Absorption and Identification invert a single resolution step. The rules of Inter- and Intra-Construction introduce “predicate invention,” which leads to reducing the hypothesis space and the length of clauses. 
	
	\textbf{Inverse Implication}. Since the deductive inference rule is incomplete regarding implication among clauses, extensions of inductive inference under $\theta$ - subsumption have been studied under the header “Inverting Implication.” The inability to invert implication between clauses limits the completeness of Inverse Resolution and RLGGs since $\theta$ - subsumption is used in place of clause implication in both. The difference between $\theta$ - subsumption and implication between clauses $C$ and $D$ are only pertinent when $C$ can self-resolve. Attempts were made to a) extend Inverse Resolution, and b) use a mixture of Inverse Resolution and LGG \cite{idestam1993generalization} to solve the problem. The extended inverse resolution method suffers from problems of non-determinacy. Due to the non-determinacy problem, The development of Algorithms regarding Inverse Implication in ILP is limited. Idestam-Almquist's use of LGG suffers from the standard problem of intractably large clauses. Both approaches are incomplete for inverting implication, though Idestam-Almquist's technique is complete for a restricted form of entailment called T-implication \cite{idestam1992learning}.
	
	\textbf{Inverse Entailment}. The general problem specification of ILP is, given background knowledge $B$ and examples $E$ find the simplest consistent hypothesis $H$ such that $B \wedge H \models E$. In general, $B$, $H$, and $E$ could be arbitrary logic programs. Each clause in the simplest $H$ should explain at least one example since there is a simpler $H'$ that will do otherwise. Consider then the case of $H$ and $E$, each being single Horn clauses. By rearranging entailment relationship $B \wedge {H} \models {E}$, Muggleton \cite{muggleton1995inverse} proposed the ``Inverse Entailment": 
	
	$$B \wedge \neg{E} \models \neg{H}$$,
	
	particularly, $\neg{\bot}$ is the (potentially infinite) conjunction of ground literals which are true in all models of $B \wedge \neg{E}$. Since $H$ must be true in every model of $B \wedge \neg{E}$, it must contain a subset of the ground literals in $\neg{\bot}$. Therefore    
	$$B \wedge \neg{E} \models \neg{\bot} \models \neg{H}$$
	$$H \models \bot$$
	A subset of the solutions for $H$ can be found by considering the clauses which
	$\theta$ - subsume $\bot$. The complete set of candidates for $H$ can be found by considering all clauses which $\theta$ - subsume sub-saturants of $\bot$.
	
	\subsection{ILP systems}
\begin{figure*}[h]
		\centering
		\captionsetup{justification=centering}
		\includegraphics[width=0.9\textwidth,height=1.1in]{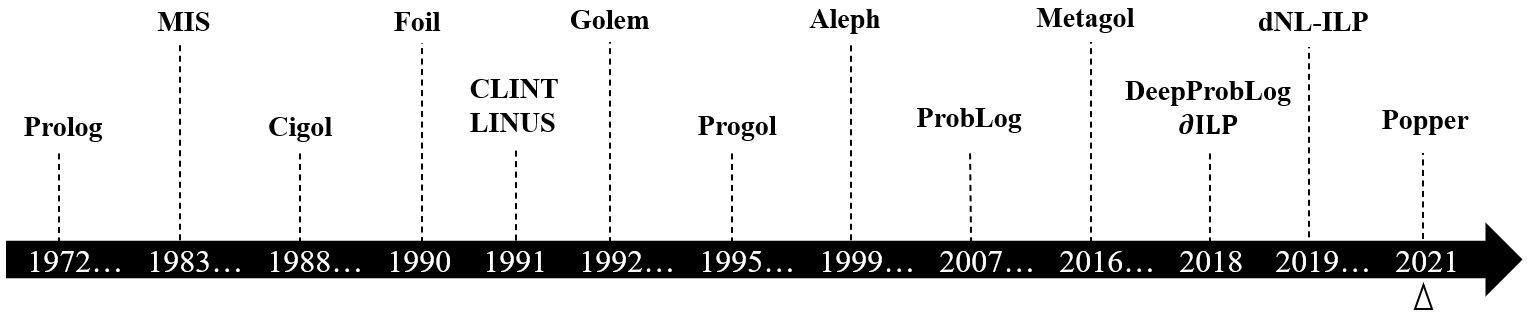}
		\caption{Logic learning systems timeline}
		\label{fig:ilp_timeline}
	\end{figure*}

	A large amount of logic learning systems have been developed based on the inference rules we mentioned before. Fig. \ref{fig:ilp_timeline} shows the timeline of the development of logic learning systems, including Prolog \cite{bratko2001prolog}, MIS \cite{shapiro1982algorithmic}, CLINT \cite{de1991clint}, Foil \cite{quinlan1990learning}, LINUS \cite{lavravc1991learning}, Golem \cite{muggleton1990efficient}, Aleph \cite{srinivasanaleph}, Progol \cite{muggleton1995inverse}, Cigol \cite{muggleton1988machine}, Metagol \cite{metagol}, ProbLog \cite{de2007problog}, DeepProbLog \cite{manhaeve2018deepproblog}, $\partial$ILP \cite{evans2018learning}, and Popper \cite{cropper2021learning}. Note that there are no common-used inverse implication systems because it suffers from problems of non-determinacy.
	
%% file: sec030405.tex
	\section{Variants of ILP}  
	\label{sec:variants-of-ilp}
	Traditional ILP frameworks that we have discussed in section \ref{sec:background} mainly have three limitations: most ILP systems 1) can not address noisy data, 2) can not deal with predicates invention directly and recursion effectively, and 3) can not learn hypothesis $H$ efficiently due to the massive hypothesis space. Variants of ILP systems have been developed to solve the problem mentioned above. Probabilistic ILP \cite{de2008probabilistic} becomes a powerful tool when dealing explicitly with uncertainty, Meta-interpretive learning \cite{muggleton2014meta} holds merits on predicate invention and recursive generalizations, and differentiable ILP \cite{evans2018learning} can speed up the learning process and also robust to noise and error. We will introduce each of them in this section.

	\subsection{Probabilistic inductive logic programming}
	Probabilistic inductive logic programming (PILP) is a machine learning technique based on probabilistic logic programming. It addresses one of the central questions of artificial intelligence by integrating probabilistic reasoning with machine learning, and first-order relational logic representations \cite{de2008probabilistic}. Dealing explicitly with uncertainty makes PILP more powerful than ILP and, in turn, than traditional attribute-value approaches \cite{de2008probabilistic}. It also provides better predictive accuracy and understanding of domains and has become a growth path in the machine learning community.
	
	The terms used in PILP are close to those in ILP with small differences: since negative examples conflict with the usual view on learning examples in statistical learning (the probability of a failure is zero), the definition of PILP problem uses observed and unobserved examples instead of positive and negative ones:
	
	\textbf{PILP problem.} Given a set $E = E_{p} \cup E_{i}$ of observed and unobserved examples $E_{p}$ and $E_{i}$ (with $E_{p} \cap E_{i}= \emptyset$) over some example language $L_{E}$, a probabilistic covers relation covers$(e, H, B) = P(e | H, B)$, a probabilistic logical language $L_{H}$ for hypotheses, and a background theory $B$, find a hypothesis $H^{*}$ in $L_{H}$ such that: $$H^{*}=argmax_{H}score(E, H, B)$$ and the following constraints hold: $$\exists e_{p} \in E_{p}: covers(e_{p}, H^{*}, B)>0$$ $$\exists e_{i} \in E_{i}: covers(e_{i}, H^{*}, B)=0$$ The score is some objective function,usually involving the probabilistic covers relation of the observed examples such as the observed likelihood $\prod_{e \in E} covers(e_{p}, H^{*}, B)$ \cite{de2008probabilistic}.
	We denote $H=(L, \lambda)$, where $L$ represents all first-order logic program rules in $H$, and $\lambda$ indicates probabilistic parameters. Two subtasks should be considered when solving probabilistic ILP learning problems: 1) Parameter estimation, where it is assumed that the underlying logic program $L$ is fixed, and the learning task consists of estimating the parameters $\lambda$ that maximize the likelihood. 2) Structure learning where both $L$ and $\lambda$ have to be learned from the data.
	
	Similar to the ILP learning problem, the language $L_{E}$ is selected for representing the examples, and the probabilistic covers relation determines different learning settings. In ILP, this leads to learning from interpretations \cite{de1994first}, from proofs \cite{de2008probabilistic}, and from entailment \cite{plotkin1970note}. Therefore, it should be no surprise that this very same distinction also applies to probabilistic knowledge representation formalisms. There are three PILP settings as well: Probabilistic Learning from Interpretations, from Entailment, and proofs \cite{de2008probabilistic}. The main idea is to lift ILP settings by associating probabilistic information with clauses and interpretations and by replacing ILP’s deterministic covers relation with a probabilistic one. The large majority of PILP techniques proposed so far fall into the learning from interpretations setting, including parameter estimation of probabilistic logic programs \cite{koller1997learning}, learning of probabilistic relational models \cite{getoor2002learning}, parameter estimation of relational Markov models \cite{taskar2012discriminative}, learning of object-oriented Bayesian networks \cite{langseth2001structural}, learning relational dependency networks \cite{neville2004dependency}, and learning logic programs with annotated disjunctions \cite{riguzzi2004learning}. To define probabilities on proofs, ICL \cite{poole1997independent}, Prism \cite{sato1997prism}, and stochastic logic programs \cite{muggleton1996stochastic} attach probabilities to facts (respectively clauses) and treat them as stochastic choices within resolution. PILP techniques that learn from proofs have been developed, including Hidden Markov model induction by Bayesian model merging \cite{stolcke1993hidden}, relational Markov models \cite{anderson2002relational}, and logical hidden Markov models \cite{kersting2006logical}. Learning from interpretations setting has been investigated for learning stochastic logic programs \cite{cussens2001parameter}, \cite{muggleton2000learning} and for parameter estimation of Prism programs \cite{kameya2004yet}, \cite{sato2001parameter} from observed examples.
	
	Many algorithms have been developed since the PILP framework: Muggleton describes a method, which based on an approximate Bayes ``Maximum A Posterior probability" (MAP) algorithm, for learning Stochastic Logic Programs (SLPs) from examples and background knowledge \cite{muggleton2000learning}, which is considered one of the earliest structure learning method for PILP; De Raedt et al. upgrade rule learning to a probabilistic setting, in which both the examples themselves as well as their classification can be probabilistic \cite{de2010probabilistic}; To solve ``large groundings” problem, William Wang et al. present a first-order probabilistic language which is well-suited to approximate ``local" grounding \cite{wang2013programming}; The algorithm ``Structure LearnIng of ProbabilistiC logic progrAmS with Em over bdds" (SLIPCASE) performs a beam search in the space of the language of Logic Programs with Annotated Disjunctions (LPAD) using the log-likelihood of the data as the guiding heuristics \cite{bellodi2011learning}; Fierens et al. investigate how classical inference and learning tasks known from the graphical model community can be tackled for probabilistic logic programs \cite{fierens2015inference}; and the algorithm ``Structure LearnIng of Probabilistic logic programs by searching OVER the clause space" (SLIPCOVER) performs a beam search in the space of probabilistic clauses and a greedy search in the space of theories, using the log-likelihood of the data as the guiding heuristics \cite{bellodi2015structure}.

	\subsection{Meta-interpretive learning}
	
	Meta-interpretive learning (MIL) is a framework developed by Muggleton et al., which uses higher-order metarules to support predicate invention and learning of recursive definitions \cite{cropper2015logical}. Metarules, second-order Horn clauses, are widely discussed \cite{cropper2019learning}, \cite{cropper2015logical}, \cite{morel2019typed}, \cite{albarghouthi2017constraint} as a form of declarative bias. Metarules define the structure of learnable programs, which in turn defines the hypothesis space. For instance, to learn the $grandparent/2$ relation given the $parent/2$ relation, the $chain$ metarule \footnote{Common-used metarules \cite{metagol}: 1) Identity: $P(A,B)\leftarrow Q(A,B)$; 2) Inverse: $P(A,B)\leftarrow Q(B,A)$; 3) precon: $P(A,B)\leftarrow Q(A),R(A,B)$; 4) postcon: $P(A,B)\leftarrow Q(A,B),R(B)$; 5) chain: $P(A,B)\leftarrow Q(A,C),R(C,B)$; 6) recursive: $P(A,B)\leftarrow Q(A,C),P(C,B)$.} would be suitable:
	$$P(A,B)\leftarrow Q(A,C),R(C,B)$$
	In this metarule, the letters $P$, $Q$, and $R$ denote existentially quantified second-order variables (variables that can be bound to predicate symbols), and the letters $A$, $B$ and $C$ denote universally quantified first-order variables (variables that can be bound to constant symbols). Given the chain metarule, the background $parent/2$ relation, and examples of the $grandparent/2$ relation, the learner will try to find the correct substitutions for predicates, one of the correct solutions would be: $$\{P/grandparent, Q/parent, R/parent\} $$
	and the substitution result is: 
	$$grandparent(A,B)\leftarrow parent(A,C),parent(C,B)$$
	We will discuss the MIL problem after the description of metarules. Before that, we define what MIL input is: An MIL input is a tuple ($B,E^+,E^-,M$), where $B$ is a set of Horn clauses denoting background knowledge, $E^+$ and $E^-$ are disjoint sets of ground atoms representing positive and negative examples respectively, and M is a set of metarules. An MIL problem \cite{cropper2019logical} can be defined from an MIL input:
	\\
	\noindent Given a MIL input ($B,E^+,E^-,M$), the MIL problem is to return a logic program hypothesis $H$ such that:
	\begin{itemize}
		\item $ \forall c\in H,\exists m\in M$ such that $c=m\theta$, where $\theta$ is a substitution that grounds all the existentially quantified variables in $m$.
		\item $H\cup B \vDash E^+$
		\item $H\cup B \nvDash E^-$
	\end{itemize}
	$H$ can be considered as a solution to the MIL problem. Based on the equation $c=m\theta$, MIL focuses on searching for $\theta$ instead of $H$ through abductive reasoning in second-order logic.
	
	As the $grandparent$ task shown before, MIL could be considered as an FOL rule searching problem based on metarules. Since any first-order predicates could substitute second-order variables, the searching of a logic program hypothesis $H$ is more flexible than that of ILP: if second-order variables are replaced by those predicates that do not exist in B, new predicates will be invented (predicate invention); if they are replaced by the same predicates in both head and body of the metarules, the recursive definition will be learned. 
	
	As we discussed before, the metarules determine the structure of permissible rules, which in turn defines the hypothesis space. Deciding which metarules to use for a given learning task is a major open problem and is a trade-off between efficiency and expressivity: we wish to use fewer metarules as the hypothesis space grows given more metarules, but if we use too few metarules, we will lose expressivity. Also, the hypothesis space of MIL highly depends on the metarules we choose. For example, The $identity$ metarule $P(A, B)\leftarrow Q(A, B)$ can not be learned from the $chain$ metarule $P(A, B)\leftarrow Q(A, C), R(C, B)$. Cropper and Muggleton \cite{cropper2015logical} demonstrate that irreducible or minimal sets of metarules can be found automatically by applying Plotkin's clausal theory reduction algorithm. When this approach is applied to a set of metarules consisting of an enumeration of all metarules in a given finite hypothesis language, they show that, in some cases, as few as two metarules are complete and sufficient for generating all hypotheses. Nevertheless, for expected hypothesis $H^{*}$, the learned model $H^{'}$ is only equivalence with the actual model $H^{*}$ semantically, not literally ($H^{'}$ usually contains more hypothesis than $H^{*}$), so reducing metarules does not necessarily improve the learning efficiency of MIL.
	
	Advances have been made to increase the performance of MIL recently. A new reduction technique, called derivation reduction \cite{cropper2019logical}, has been introduced to find a finite subset of a Horn theory from which the whole theory can be derived using SLD-resolution. Recent work \cite{morel2019typed} also shows that adding types to MIL can improve learning performance, and type checking can reduce the MIL hypothesis space by a cubic factor, which can substantially reduce learning times. Extended MIL \cite{cropper2019learning} has also been used to support learning higher-order programs by allowing for higher-order definitions to be used as background knowledge. Recently, Popper \cite{cropper2021learning} supports types, learning optimal solutions, learning recursive programs, reasoning about lists and infinite domains, and hypothesis constraints by combining ASP and Prolog.
	
	\subsection{Differentiable inductive logic programming}
	\label{subsec:differentiable-inductive-logic-programming}
	Differentiable inductive logic programming ($\partial$ILP) that combines intuitive perceptual with conceptually interpretable reasoning has several advantages: robust to noise and error, data-efficient, and produces interpretable rules \cite{evans2018learning}. $\partial$ILP implements differentiable deduction over continuous values. The gradient of the loss concerning the rule weights, which we use to minimize classification loss, implements a continuous form of induction \cite{evans2018learning}.
	
	The set $P$ (positive examples) and $N$ (negative examples) are used to form a new set $\Lambda$:
	$$\Lambda=\{(\gamma,1)|\gamma\in P\}\cup\{(\gamma,0)|\gamma\in N\}$$
	Each pair $(\gamma,\lambda)$ indicates that whether the atom $\gamma$ is in $P$ ($\lambda=1$) or $N$ ($\lambda=0$).    A differentiable model is constructed to implement the conditional probability
	of $\lambda$ for a ground atom $\alpha$:
	$$p(\lambda|\alpha,W,\Pi,L,B)$$
	$W$ is a set of clause weights, $\Pi$ is a program template, $L$ is a language frame, and $B$ is a set of background assumptions. The goal is to match the predicted label $p(\lambda|\alpha,W,\Pi,L,B)$ with the actual label $\lambda$ in the pair $(\gamma,\lambda)$. Minimizing the expected negative log-likelihood is performed:
	
	\begin{equation}\nonumber
	\begin{split}
	loss=-E_{(\gamma,\lambda)\sim\Lambda}[\lambda\cdot \log p(\lambda|\alpha,W,\Pi,L,B)+\\(1-\lambda)\cdot \log (1-p(\lambda|\alpha,W,\Pi,L,B))]
	\end{split}
	\end{equation}
	
	The conditional probability $p(\lambda|\alpha,W,\Pi,L,B)$ is calculated by four functions: $f_{extract}$, $f_{infer}$, $f_{convert}$, and $f_{generate}$.
	
	\begin{equation}\nonumber
	\begin{split}
	p(\lambda|\alpha,W,\Pi,L,B)=f_{extract}(f_{infer}(f_{convert}(B),\\f_{generate}(\Pi,L),W,T),\alpha)
	\end{split}
	\end{equation}
	
	The $f_{extract}$ extracts the value from an atom, the $f_{convert}$ converts the elements of $B$ to 1 and others to 0, and the $f_{generate}$ generates a set of clauses from $\Pi$ and $L$. The $f_{infer}$ performs $T$ steps of forward-chaining inference where $T$ is part of $\Pi$.
	To show how inference is performed over multiple time steps, each clause $c$ has been translated into a function $F_{c}:[0,1]^{n}\rightarrow [0,1]^{n}$ on valuations \cite{evans2018learning}. Each intensional predicate $p$ is defined by two clauses generated from two rule templates $\tau_{p}^{1}$ and $\tau_{p}^{2}$ \footnote{Rule template $\tau$ describes a range of clauses that can be generated.}. $G_{p}^{j,k}$ \footnote{$G_{p}^{j,k}$ combines the application of two functions $F_{p}^{1,j}$ and $F_{p}^{2,k}$. $F_{p}^{i,j}$ is the valuation function corresponding to the clause $C_{p}^{i,j}$} is the result of applying $C_{p}^{1,j}$ \footnote{$C_{p}^{i,j}$ is the j'th clause of the i'th rule template $\tau_{p}^{i}$ for intensional predicate $p$.} and $C_{p}^{2,k}$ and taking the element-wise max \cite{evans2018learning}:\\
	$$G_{p}^{j,k}(a)=x$$ where $$x[i]=\max(F_{p}^{1,j}(a)[i],F_{p}^{2,k}(a)[i])$$
	$a_{t}$ depicts conclusions after $t$ time-steps of inference. $a_{0}[x]$ equals to 1 if $\gamma$ belongs to $B$, and 0 otherwise. Intuitively, $c_{t}^{p,j,k}$, which equals to $G_{p}^{j,k}(a_{t})$, is the result of applying one step of forward chaining inference to at using clauses $C_{p}^{1,j}$ and $C_{p}^{2,k}$. The weighted average of the $c_{t}^{p,j,k}$ can be defined using the softmax of the weights:
	$$b_{t}^{p}=\sum_{j,k}c_{t}^{p,j,k}\cdot \frac{e^{W_{p}[j,k]}}{\sum_{j^{'},k^{'}}e^{W_{p}[j^{'},k^{'}]}}$$
	and the successor $a_{t+1}$ of $a_{t}$ is the probabilistic sum:
	$$a_{t+1}=a_{t}+\sum_{p\in P_{i}}b_{t}^{p}-a_{t}\cdot \sum_{p\in P_{i}}b_{t}^{p}$$
	Besides $\partial$ILP, several neural program synthesis models have been used to produce explicit human-readable programs: RobustFill \cite{devlin2017robustfill} performs an end-to-end synthesis of programs from examples by using a modified attention RNN to allow encoding of variable-sized sets of I/O pairs; a Differentiable Forth Interpreter \cite{bovsnjak2017programming} is an end-to-end interpreter for the programming language Forth which enables programmers to write program sketches with slots that can be filled with behavior trained from program input-output data; Neural Theorem Provers (NTPs) \cite{rocktaschel2017end}, end-to-end differentiable provers for basic theorems formulate as queries to a knowledge base, and use Prolog’s backward chaining algorithm as a recipe for recursively constructing neural networks that are capable of proving queries to a knowledge base using subsymbolic representations. Payani et al. propose a novel paradigm, called Differentiable Neural Logic ILP (dNL-ILP), for solving ILP problems via deep recurrent neural networks  \cite{payani2019inductive}. The dNL-ILP, in contrast to the majority of past methods, directly learns the symbolic logical predicate rules instead of searching through the space of possible first-order logic rules by using some restrictive rule templates.
	
	\section{Symbolic-based integration}
	\label{sec:data-driven-predicate-extraction}
	ILPs hold certain merits on explainability due to their symbolic nature and also show drawbacks such as lack of robustness, cannot deal with big data, and being hard to express complex non-linear decision surfaces. With the development of AI in other fields, the integration of symbolic AI and other ML methods can make use of the strengths and avoid weaknesses of each method. We present the two most important integrations in this section: Statistical relational learning (SRL) and Neural-symbolic AI (NeSy).
	
	\subsection{Statistical relational learning}
	\label{sec:statistical-relational-learning}
	
	Most machine learning algorithms assume that the data is independently distributed. In the real world, objects have different kinds of relations between them, which means that the objects in the dataset have relations to each other, have different types, and have multiple kinds of distribution. Statistical relational learning (SRL) is a subdiscipline of artificial intelligence and machine learning that is concerned with domain models that exhibit both uncertainties (which can be dealt with using statistical methods) and complex, relational structures \cite{koller2007introduction}, \cite{rossi2012transforming}.
	
	SRL usually provides not only a better understanding of domains and predictive accuracy but a more complex learning and inference process. The difference between PILP and SRL is that SRL has started from a statistical and probabilistic learning perspective and extended probabilistic formalisms with relational aspects, while PILP takes a different perspective, starting from ILP. As we mentioned before, SRL focuses on learning when samples are non-i.i.d. Domains where data is non-i.i.d. are widespread; examples include web search, information extraction, perception, medical diagnosis/epidemiology, molecular and systems biology, social science, security, ubiquitous computing, etc. In all of these domains, modeling dependencies between examples can significantly improve predictive performance and lead to a better understanding of the relevant phenomena. The knowledge representation formalisms developed in SRL use first-order logic to describe relational properties of a domain in a general manner (universal quantification) and draw upon probabilistic graphical models such as Bayesian networks or Markov networks to model the uncertainty, while others also build upon the methods of ILP.
	
	Probabilistic relational models (PRMs)\cite{koller1999probabilistic} are a rich representation language for structured statistical models, such as a probabilistic graphical model \cite{chen2008learning}. A PRM models the uncertainty over the attributes of objects in the domain and uncertainty over the relations between the objects. The model specifies, for each attribute of an object, its (probabilistic) dependence on other attributes of that object and attributes of related objects. Typical PRMs, including Relational Bayesian networks (RBN) \cite{jaeger1997relational}, Relational Markov networks (RMN) \cite{taskar2007relational}, and Relational Dependency Networks (RDN) \cite{neville2007relational} can deal with non-complete and non-accurate relational data. However, learning a graphical model requires structure learning and parametric learning. Structure learning is a combinatorial optimization problem that has high complexity. Besides, due to the convergence for parameter learning of RMNs and RDNs being slow, some approximation strategies are commonly used. Therefore, the PRMs are suitable for processing small-scale data.
	
	Probabilistic logic models (PLMs) \cite{chen2008learning} that combine probabilities with first-order logic can handle relational data well. Typical PLMs include probabilistic Horn abduction (PHA) \cite{poole1993probabilistic}, Bayesian logic programming (BLP) \cite{kersting20071}, and Markov logic networks (MLNs) \cite{richardson2006markov}. The learning speed of PRMs is slow since PLMs are based on graph models, so they are only suitable for processing a small amount of data. PLMs can handle noisy data and provide better predictive accuracy and understanding of domains but suffer from the computational complexity of inference.
	
	\subsection{Neural-Symbolic AI}
	Neural Networks can deal with mislabelled and noisy data but suffer from problems such as data-efficiency and explainability. Symbolic methods are data-efficient along with a number of drawbacks, such as a lack of handling error data and problems in the data. Blending deep neural network topologies with symbolic reasoning techniques creates a more advanced version of AI, i.e., Neural-Symbolic AI (NeSy): the hybrid model aims to combine robust learning in neural networks with reasoning and explainability via symbolic representations for network models \cite{garcez2020neurosymbolic}.
    
    To bring together neural networks and symbolic AI, Henry Kautz indicates a taxonomy including five different types of neural-symbolic AI at AAAI-2020 \cite{kautz2020}. In Kautz's taxonomy, Type 1 is a standard deep learning procedure in which the input and output of a neural network can be made of vectors of symbols, e.g., text in the case of classification, entity extraction, and translation. Type 2 is a neural pattern recognition subroutine within a symbolic problems solver such as Monte Carlo tree search, e.g., AlphaGo and self-driving car paradigms. Type 3 is a hybrid system that takes symbolic rules, e.g., $A\leftarrow B$, as an input-output training pair $(A, B)$, i.e., the knowledge of symbolic rules will be learned by the neural network. One example uses integration to solve differential equations \cite{lample2019deep}. Type 4 cascades from the neural network into symbolic reasoner, i.e., the neuro-symbolic concept learner (NS-CL) \cite{mao2019neuro} and Conditional Theorem Provers (CTPs) \cite{minervini2020learning}. Type 5 embeds symbolic reasoning inside a neural engine, e.g., in business AI, when attention to concepts is very high, they are decoded into symbolic entities in an attention schema. A goal that appears in the attention schema indicates that deliberative symbolic reasoning should be initiated.
    
    Table \ref{table:papers_based_on_Kautz_categories} summarizes models for the types based on Kautz's categorization. We now briefly discuss state-of-the-art neural-symbolic models for each type of integration. Note that there are no papers fully regarding type 5. For type 1, Symbolic Deep Reinforcement Learning (SDRL) \cite{lyu2019sdrl} framework conducts high-level symbolic planning based on intrinsic goals using explicitly represented symbolic knowledge and utilizes Deep Reinforcement Learning (DRL) to learn low-level control policy, leading to improved task-level interpretability for DRL and data-efficiency; Neural-Symbolic Stack Machine (NeSS) \cite{chen2020compositional} is a differentiable neural network that operates a symbolic stack machine that supports general-purpose sequence-to-sequence generalization, to accomplish compositional generalization; LENSR \cite{xie2019embedding} is a novel approach for improving the performance of deep models by leveraging prior symbolic knowledge. For type 2, Generative Neurosymbolic Machine (GNM) \cite{jiang2020generative} combines the advantages of distributed and symbolic representation in generative latent variable models; GraIL \cite{teru2020inductive} is a GNN-based framework that predicts relations between nodes that were unseen during training and produces state-of-the-art performances in this inductive setting; a sparse-matrix reified knowledge base (KB) \cite{cohen2020scalable} is a revolutionary technique of representing a symbolic KB, which enables neural modules that are fully differentiable, faithful to the original semantics of the KB, expressive enough to model multi-hop inferences, and scalable enough to utilize with realistically large KBs. For type 3, Neural Symbolic Reader (NeRd) \cite{chen2019neural} is a scalable integration of distributed representations and symbolic operations for reading comprehension that consists of a reader that encodes text into vector representations and a programmer that generates programs, which will be executed to produce the answer; Know-Evolve \cite{trivedi2017know} is a novel deep evolutionary knowledge network that learns non-linearly evolving entity representations over time; neural equivalence networks (EQNETs) \cite{allamanis2017learning} focus on the challenge of learning continuous semantic representations of algebraic and logical expressions. For type 4, Neuro-Symbolic Concept Learner (NS-CL) \cite{mao2019neuro} learns by looking at images and reading associated questions and answers, without any explicit supervision such as class labels for objects; CTP \cite{minervini2020learning} is an extension to NTP that uses gradient-based optimization to learn the best rule selection technique; DrKIT \cite{dhingra2020differentiable} is a differentiable module that is capable of answering multi-hop questions directly using a large entity-linked text corpus.
    \begin{table}[]
        \centering
        \begin{tabular}{|c|c|c|}
        \hline
        \textbf{Type} & \textbf{Category}                    & \textbf{Models}                                                                   \\ \hline
        1             & {[}Symbolic Neuro Symbolic{]}        & \begin{tabular}[c]{@{}c@{}}SDRL, NeSS,\\ LENSR\end{tabular}                       \\ \hline
        2             & {[}Symbolic{[}Neuro{]}{]}            & \begin{tabular}[c]{@{}c@{}}GNM, GraIL, \\ Sparsematrix \\ reified KB\end{tabular} \\ \hline
        3             & {[}Neuro $\cup$ compile(Symbolic){]} & \begin{tabular}[c]{@{}c@{}}NeRd, EQNET\\ Know-Evolve\end{tabular}                 \\ \hline
        4             & {[}Neuro $\rightarrow$ Symbolic{]}    & \begin{tabular}[c]{@{}c@{}}DrDKIT,\\ NS-CL, CTP\end{tabular}   \\ \hline
        5             & {[}Neuro{[}Symbolic{]}{]}            & N/A                                                                               \\ \hline
        \end{tabular}
        \caption{Papers based on Kautz's taxonomy.}
        \label{table:papers_based_on_Kautz_categories}
    \end{table}
    
    \subsection{SRL vs. NeSy}
    Although SRL and NeSy are merging symbolic reasoning with a different fundamental learning paradigm: SRL is the integration of logic and probability, while NeSy is the combination of symbolic reasoning and neural networks. They have a lot in common, and there are interactions between these two fields. De Raedt et al. identify seven dimensions that SRL and NeSy approaches have in common \cite{de2020statistical}. These seven dimensions are: 1) type of logic, 2) model vs. proof-based inference, 3) directed vs. undirected models, 4) logical semantics, 5) learning parameters or structure, 6) representing entities as symbols or sub-symbols, and 7) integrating logic with probability and/or neural computation.
    
    We add another dimension, SRL, to represent multiple SRL and NeSy models in Table 2 in addition to summarizing numerous NeSy structures in \cite{de2020statistical}, along the seven dimensions. Note that, in dimension 1, logic programming and FOL are mostly used for SRL, while models adapt to different types of logic based on the complexity and expressiveness of logic for NeSy. In dimension 4, probabilistic Boolean logic is the most common semantic in SRL. Whereas in NeSy, many probabilistic approaches use neural components to parameterize the underlying distribution, while fuzzy logic is used for computational reasons mostly and
    to relax rules. In dimension 6, for NeSy, the input and intermediate representations are sub-symbolic, while the output representations can be either symbolic or sub-symbolic; all representations in SRL are symbolic.

    \begin{table}[]
    \centering
        \begin{tabular}{|cc|c|cc|}
        \hline
        \multicolumn{2}{|c|}{}                                                                                                                                                                 & \textbf{SRL}                                                        & \multicolumn{2}{c|}{\textbf{NeSy}}                                                                                                                                  \\ \hline
        \multicolumn{1}{|c|}{\multirow{4}{*}{\textbf{\begin{tabular}[c]{@{}c@{}}Dimension 1\\ (P)ropositional\\ (R)elational\\ (FOL)\\ (L)ogic-\\ (P)rogramming\end{tabular}}}} & \textbf{P}   & \begin{tabular}[c]{@{}c@{}}ProbLog \cite{de2007problog},\\ Prism \cite{sato1997prism}\end{tabular}            & \multicolumn{2}{c|}{SL \cite{xu2018semantic}}                                                                                                                                             \\ \cline{2-5} 
        \multicolumn{1}{|c|}{}                                                                                                                                                  & \textbf{R}   & -                                                            & \multicolumn{2}{c|}{\begin{tabular}[c]{@{}c@{}}$\partial$ILP \cite{evans2018learning},\\ NLM \cite{dong2019neural}\end{tabular}}                                                                                   \\ \cline{2-5} 
        \multicolumn{1}{|c|}{}                                                                                                                                                  & \textbf{FOL} & MLN \cite{richardson2006markov}                                                                & \multicolumn{2}{c|}{\begin{tabular}[c]{@{}c@{}}LTN \cite{serafini2016logic},\\ RNM \cite{marra2020relational}\end{tabular}}                                                                                             \\ \cline{2-5} 
        \multicolumn{1}{|c|}{}                                                                                                                                                  & \textbf{LP}  & -                                                            & \multicolumn{2}{c|}{\begin{tabular}[c]{@{}c@{}}DeepProbLog \cite{manhaeve2018deepproblog},\\ NLProlog \cite{weber2019nlprolog}\end{tabular}}                                                                                       \\ \hline
        \multicolumn{1}{|c|}{\multirow{2}{*}{\textbf{\begin{tabular}[c]{@{}c@{}}Dimension 2\\ (M)odel-based\\ (P)roof-based\end{tabular}}}}                                     & \textbf{M}   & \begin{tabular}[c]{@{}c@{}}MLN,\\ PRM \cite{koller1999probabilistic}\end{tabular}                  & \multicolumn{2}{c|}{\begin{tabular}[c]{@{}c@{}}LTN, \\ RNM\end{tabular}}                                                                                            \\ \cline{2-5} 
        \multicolumn{1}{|c|}{}                                                                                                                                                  & \textbf{P}   & \begin{tabular}[c]{@{}c@{}}PLP \cite{de2015probabilistic},\\ SLP \cite{muggleton1996stochastic}\end{tabular}                  & \multicolumn{2}{c|}{\begin{tabular}[c]{@{}c@{}}$\partial$ILP,\\ DeepProbLog\end{tabular}}                                                                           \\ \hline
        \multicolumn{1}{|c|}{\multirow{2}{*}{\textbf{\begin{tabular}[c]{@{}c@{}}Dimension 3\\ (D)irected\\ (U)ndirected\end{tabular}}}}                                         & \textbf{D}   & \begin{tabular}[c]{@{}c@{}}PRM,\\ PLP\end{tabular}                  & \multicolumn{2}{c|}{\begin{tabular}[c]{@{}c@{}}$\partial$ILP,\\ NLM\end{tabular}}                                                                                   \\ \cline{2-5} 
        \multicolumn{1}{|c|}{}                                                                                                                                                  & \textbf{U}   & \begin{tabular}[c]{@{}c@{}}MLN,\\ PSL \cite{bach2015hinge}\end{tabular}                  & \multicolumn{2}{c|}{\begin{tabular}[c]{@{}c@{}}SL,\\ LTN\end{tabular}}                                                                                              \\ \hline
        \multicolumn{1}{|c|}{\multirow{3}{*}{\textbf{\begin{tabular}[c]{@{}c@{}}Dimension 4\\ (L)ogic\\ (P)robability\\ (F)uzzy\end{tabular}}}}                                 & \textbf{L}   & \begin{tabular}[c]{@{}c@{}}MLN,\\ PLP\end{tabular}                          & \multicolumn{2}{c|}{\begin{tabular}[c]{@{}c@{}}NTP \cite{rocktaschel2017end},\\ NLM\end{tabular}}                                                                                             \\ \cline{2-5} 
        \multicolumn{1}{|c|}{}                                                                                                                                                  & \textbf{F}   & PSL                                                             & \multicolumn{2}{c|}{\begin{tabular}[c]{@{}c@{}}$\partial$ILP,\\ LTN\end{tabular}}                                                                                   \\ \cline{2-5} 
        \multicolumn{1}{|c|}{}                                                                                                                                                  & \textbf{P}   & \begin{tabular}[c]{@{}c@{}}MLN,\\ PLP,\\ PSL\end{tabular}                                     & \multicolumn{2}{c|}{\begin{tabular}[c]{@{}c@{}}SL,\\ DeepProbLog\end{tabular}}                                                                                      \\ \hline
        \multicolumn{1}{|c|}{\multirow{2}{*}{\textbf{\begin{tabular}[c]{@{}c@{}}Dimension 5\\ (P)arameter\\ (S)tructure\end{tabular}}}}                                         & \textbf{P}   & \begin{tabular}[c]{@{}c@{}}ILP,\\ MLN\end{tabular}                                                                & \multicolumn{1}{c|}{\begin{tabular}[c]{@{}c@{}}LTN, \\ SL\end{tabular}}           & \multirow{2}{*}{\begin{tabular}[c]{@{}c@{}}$\partial$ILP,\\  NLM\end{tabular}}  \\ \cline{2-4}
        \multicolumn{1}{|c|}{}                                                                                                                                                  & \textbf{S}   & \begin{tabular}[c]{@{}c@{}}ILP,\\ PGM \cite{larranaga2011probabilistic}\end{tabular}                  & \multicolumn{1}{c|}{-}                                                     &                                                                                 \\ \hline
       \multicolumn{1}{|c|}{\multirow{2}{*}{\textbf{\begin{tabular}[c]{@{}c@{}}Dimension 6\\ (S)ymbolic\\ (Sub)symbolic\end{tabular}}}}                                        & \textbf{S}   & All models                                                          & \multicolumn{1}{c|}{\begin{tabular}[c]{@{}c@{}}$\partial$ILP,\\ NLM\end{tabular}} & \multirow{2}{*}{\begin{tabular}[c]{@{}c@{}}Deep-\\ ProbLog,\\  SL\end{tabular}} \\ \cline{2-4}
        \multicolumn{1}{|c|}{}                                                                                                                                                  & \textbf{Sub} & -                                                            & \multicolumn{1}{c|}{\begin{tabular}[c]{@{}c@{}}SBR,\\ LTN\end{tabular}}           &                                                                                 \\ \hline
        \multicolumn{1}{|c|}{\multirow{3}{*}{\textbf{\begin{tabular}[c]{@{}c@{}}Dimension 7 \\ (L)ogic\\ (P)robability\\ (N)eural\end{tabular}}}}                               & \textbf{L}   & \multirow{2}{*}{\begin{tabular}[c]{@{}c@{}}PLP,\\ MLN\end{tabular}} & \multicolumn{2}{c|}{\begin{tabular}[c]{@{}c@{}}$\partial$ILP,\\ RNM\end{tabular}}                                                                                   \\ \cline{2-2} \cline{4-5} 
        \multicolumn{1}{|c|}{}                                                                                                                                                  & \textbf{P}   &                                                                     & \multicolumn{2}{c|}{RNM, SL}                                                                                                                                        \\ \cline{2-5} 
        \multicolumn{1}{|c|}{}                                                                                                                                                  & \textbf{N}   & -                                                            & \multicolumn{2}{c|}{\begin{tabular}[c]{@{}c@{}}$\partial$ILP,\\ RNM\end{tabular}}                                                                                   \\ \hline
        \end{tabular}
        \caption{SRL and NeSy models along seven dimensions.}
        \label{table:srl_n_nesy}
    \end{table}
    
	\section{User-centered explainable ILP}
	\label{sec:user-centered_explainable_ilp}
    Explainable artificial intelligence (XAI) has gotten a lot of attention lately as researchers try to figure out what it means and expects. As AI systems are used by a broader and more diverse audience, researchers have realized that it is crucial to pay enough attention to AI users. From a user-centered perspective, as we surveyed ILP systems and their integrations, we briefly discuss models that could apply to ILP by answering the following questions:\\
    1) How should we create explanations for users?\\
    2) How should a user trust a model?\\
    3) How to provide explanations for symbolic-based integration, e.g., NeSy?
    
	\subsection{User-centered explanation}

    Explainability based on model users' characteristics is essential since generating explanations based on different audiences can overcome the difficulties of fulfilling all requirements simultaneously. \cite{ribera2019can} classifies users into three main groups (developers or AI researchers, domain experts, and lay users), and designs different explanations for each group with a particular purpose, content, and presents it in a specific way, which is a suitable framework for ILP systems. For developers and AI researchers, model inspection and simulation with proxy models are provided. These two types of explanations are great for validating the system, discovering flaws, and providing suggestions for improvement. Because they can understand codes, data representation structures, and statistical variances, the audience is well-suited to this mode of communication. For domain experts, who are capable of deciding when and how to question the explanation and are led to their discovery by themselves, explanations through natural language conversations or interactive visualizations are offered. For lay users, outcome explanations with several counterfactuals are used, with which users can interact to select the one most interesting to their particular case.

	\subsection{Measurement of trust}
\begin{figure*}[h]
		\centering
		\captionsetup{justification=centering}
		\includegraphics[width=0.8\textwidth,height=2.0in]{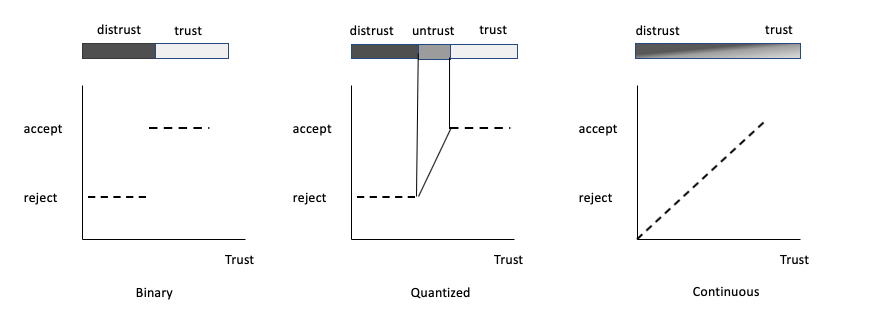}
		\caption{Trust Measurement}
		\label{fig:trust_measurement}
	\end{figure*}

	Besides creating explanations for different users, how users trust AI systems is also crucial. Yilmaz and Liu \cite{yilmaz2020model} propose a measurement of trust that is also appropriate for ILP. In this measurement, three evaluation models with increasing levels of specificity can be considered: binary evaluation, quantized/ discrete evaluation, and continuous/ spectral evaluation (Fig. \ref{fig:trust_measurement}). 
	
	\begin{table*}[h]
		\centering
		\begin{tabular}{|c|c|c|c|c|}
			\hline
			\multirow{2}{*}{\textbf{Model of Trust Evaluation}} & \multicolumn{4}{c|}{\textbf{Based on the criteria}}       \\ \cline{2-5} 
			&
			\begin{tabular}[c]{@{}c@{}}concern with\\ the problem\end{tabular} &
			\begin{tabular}[c]{@{}c@{}}ability to\\ process\\ information\end{tabular} &
			\begin{tabular}[c]{@{}c@{}}domain\\ familiarity/\\ experience/\\ expertise\end{tabular} &
			\begin{tabular}[c]{@{}c@{}}availability of\\ standards/\\ reference models\end{tabular} \\ \hline
			Binary                                              & low      & limited     & low     & lacks standards (none) \\ \hline
			Quantized                                           & moderate & moderate    & partial & some                   \\ \hline
			Spectral                                            & high     & significant & high    & substantial            \\ \hline
		\end{tabular}
		\caption{Types of Trust Evaluation}
		\label{type_trust}
	\end{table*}
	
	As shown in Fig. \ref{fig:trust_measurement}, Binary evaluation shows that model users are at the level of categorizing a model as either trusted or distrusted without any middle ground; quantized evaluation sets a threshold in the unfavorable region distinguishes trust from distrust when the model user lacks trust; continuous evaluation measures trust over a continuous range and gives model users the discretion on the degree of trust. Table \ref{type_trust} denotes how these criteria relate to the evaluation models. 
	
	\subsection{Explanation for symbolic-based integration}
	In the current state-of-the-art XAI, models that combine connectionism and symbolism are not frequently represented. The explanation for symbolic-based integration is hard since the connectionist methods in the hybrid system offer poor explainability. To conquer this problem, EXplainable Neural-Symbolic Learning \cite{diaz2022explainable} learns both symbolic and deep representations, together with an explainability metric to assess the level of alignment of machine and human-expert explanations. Campagner et al. \cite{campagner2020back} use deep learning for the automatic detection of meaningful, hand-crafted high-level symbolic features, which are then to be used by a standard and more interpretable learning model. Bennetot et al. \cite{bennetot2019towards} present a methodology based on a neural network's learning data that allows us to influence its learning and entirely fix biases while providing a fair explanation from its predictions.
	

%% file: sec060708.tex
  \section{Experiments}
	\label{sec:experiments}
	In this section, we evaluate several ILP systems following the experiment procedures in \cite{cropper2021learning} when varying: 1) optimal solution size, 2) optimal domain size, 3) example size, and 4) benchmark ILP problems. We also compare these systems from a feature perspective and summarize the comparison results. It is crucial to keep in mind that comparing ILP systems is challenging since they are typically biased and excel at distinct tasks. Thus, we do not conclude that system X is superior to system Y \cite{cropper2021learning}.
	
	\subsection{Robot} 
	\label{subsection:robots}
	This comparison aims to find the optimal solution size for each ILP system. To control the solution size, we use a strategy learning problem, i.e., \textit{Robot}: there is a robot in a $n\times n$ world. The aim is to develop a general plan for moving the robot $i$ steps toward the right in the grid from an arbitrary starting location. For example, for a $5\times 5$ world, and a start position $(1, 3)$, the goal for the robot is to move to position $(3, 3)$ within $i$ steps where $i=2$. A solution could be: $$f(A,B)\leftarrow move\_right(A,C), move\_right(C,B)$$
	
	The solution indicates that the robot could reach the final position by moving $2$ steps to the right. We fix \textit{n} to $20$ and vary \textit{i} that corresponds to the optimal solution size.
	The settings for each system are shown as follows. For Aleph, the maximum number of nodes to be searched is $50000$. For Metagol, $5$ metarules (table \ref{table:the_metarules_used_in_the_robot_experiment}) are used to guide the searching process if $i<3$, and $6$ metarules are used if $i\geq 3$ (metarules in table \ref{table:the_metarules_used_in_the_robot_experiment} and another metarule for the current step size). For Popper, there are no restrictions in the experiment.
    
    For each \textit{i} in $[1, 2, ..., 10]$, we generate $100$ positive (robot has reached the correct position within $i$ steps) and $100$ negative examples (robot has not reached the correct position within $i$ steps). If a system fails to learn a solution within a given time, it will achieve default accuracy ($50\%$). We set a timeout of $60$ seconds per task and repeat each experiment ten times to get the mean and the standard error.
	
	\begin{figure}[h]
		\includegraphics[width=0.5\textwidth,height=0.4\textwidth]{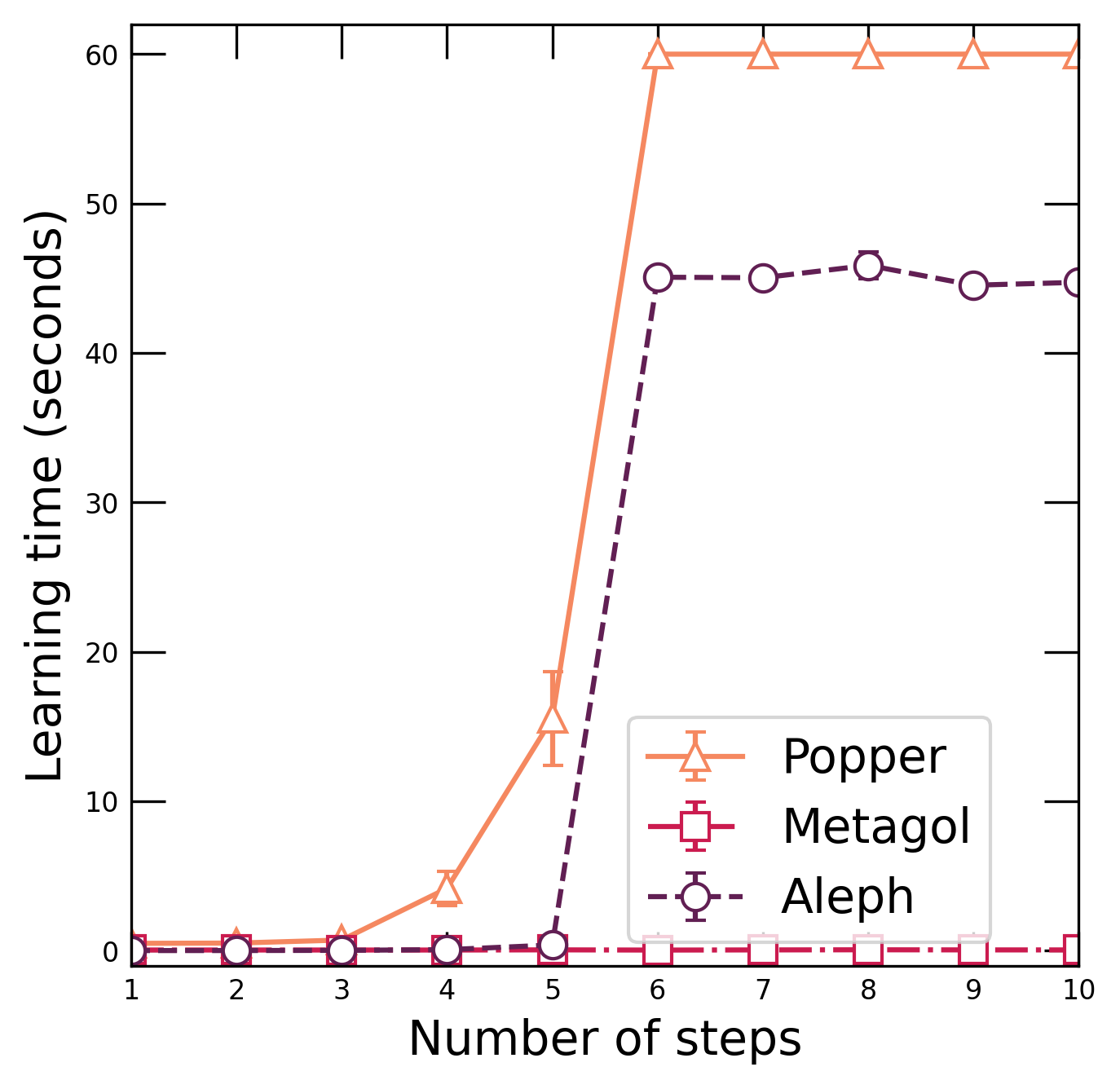}
		\caption{Learning time for $Robot$ experiment}
		\label{fig:robot_experiment}
	\end{figure}
	
	Figure \ref{fig:robot_experiment} shows that Aleph outperforms both methods when $i>4$. The learning time of Aleph and Popper increases significantly when the step size reaches $5$. The largest program Popper learns is when $i = 5$, which takes $16$ seconds, compared to less than one second for Aleph and Metagol. Popper and Aleph share comparable learning curves, while Metagol outperforms the other two because metarules lead the search efficiently. Note that the accuracies of all three systems are $100\%$ for the experiment.
	
	\subsection{Robot2}
	
	The goal of this comparison is to find the optimal domain size for each ILP system. To control the domain size, we modify the game in \ref{subsection:robots} to \textit{robot2} \cite{cropper2020inductive}: There is a robot in a $n\times n$ world. The goal is to provide a universal method for advancing the robot from any starting position to the rightmost of the grid. The objective for the robot of a $5\times 5$ world, for instance,  is to go to the location $(3, 5)$ with a start position $(3, 1)$. One solution could be: 
    
    \begin{align}
     f(A,B) & \leftarrow move\_right(A,B),at\_right(A,B) \nonumber \\
     f(A,B) & \leftarrow move\_right(A,C),f(C,B) \nonumber
    \end{align}

	The settings for all systems are described below. For Aleph, the maximum number of nodes to be searched is $50000$. For Metagol, $5$ metarules are used (table \ref{table:the_metarules_used_in_the_robot_experiment}) in the experiment. There are no restrictions for Popper.

\begin{table}[]
\centering
\begin{tabular}{l}
\hline
\begin{tabular}[c]{@{}l@{}}
$P(A,B)\leftarrow Q(A,B)$ \\ 
$P(A,B)\leftarrow Q(A,B),R(A)$ \\
$P(A,B)\leftarrow Q(A,B),R(B)$ \\
$P(A,B)\leftarrow Q(A,C),R(C,B)$ \\
$P(A,B)\leftarrow Q(A,C),P(C,B)$ \\
\end{tabular} \\ \hline
\end{tabular}
\caption{The metarules used in the robot experiment.}
\label{table:the_metarules_used_in_the_robot_experiment}
\end{table}

    We vary domain size by changing \textit{n}. For each \textit{n} in $[10, 20, ..., 100]$, we generate $100$ positive (the starting position is not at the right and the end position is at the right) and $100$ negative examples (the starting position and the end position are not at the right). If a system fails to learn a solution within a given time, it will achieve default predictive accuracy ($50\%$). We enforce a timeout of $60$ seconds per task and repeat each experiment ten times to get the mean and the standard error. 
	
	\begin{figure}[h]
		\includegraphics[width=0.5\textwidth,height=0.4\textwidth]{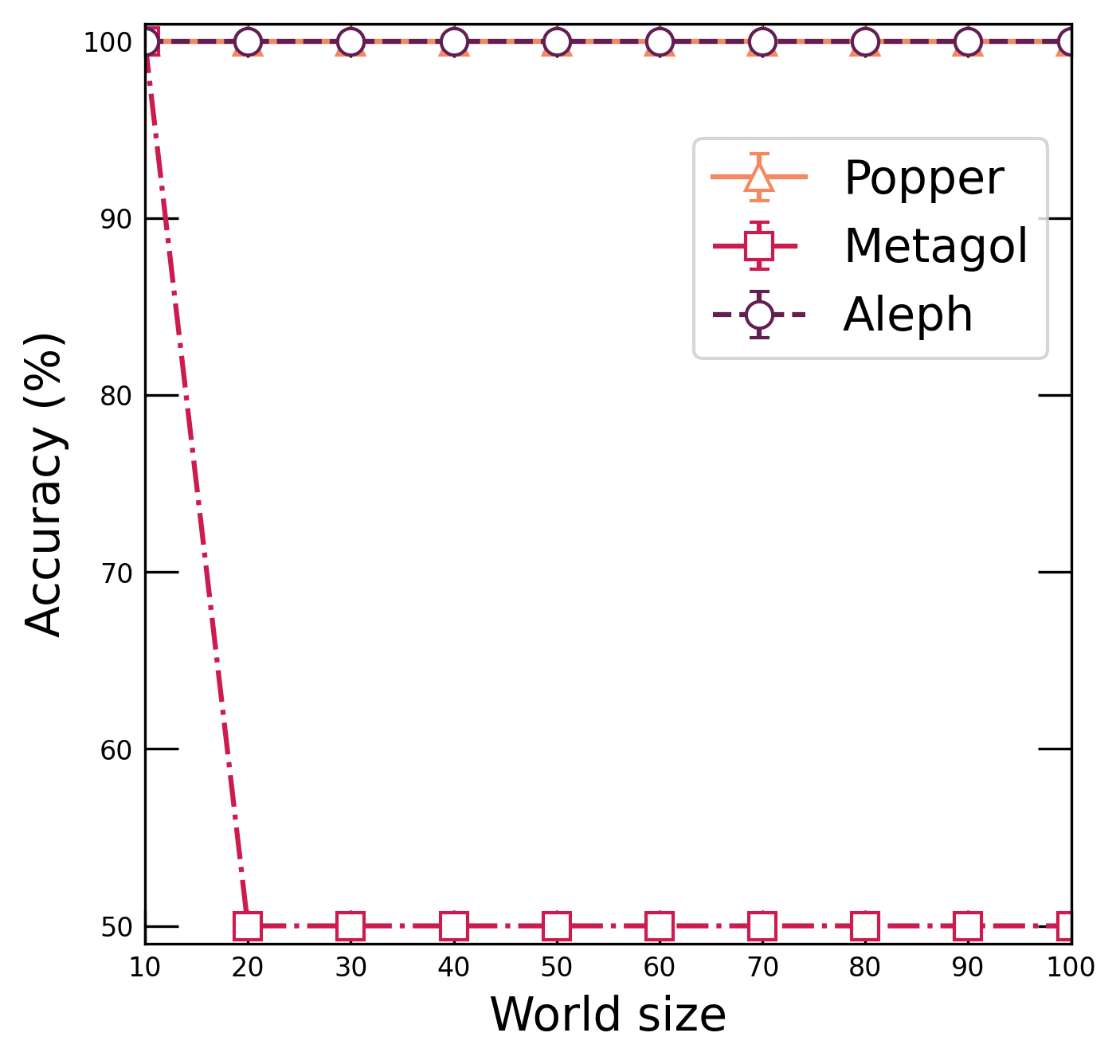}
		\caption{Accuracy for $Robot2$ experiment}
		\label{fig:robot2_accuracy}
	\end{figure}
	
	\begin{figure}[h]
		\includegraphics[width=0.5\textwidth,height=0.4\textwidth]{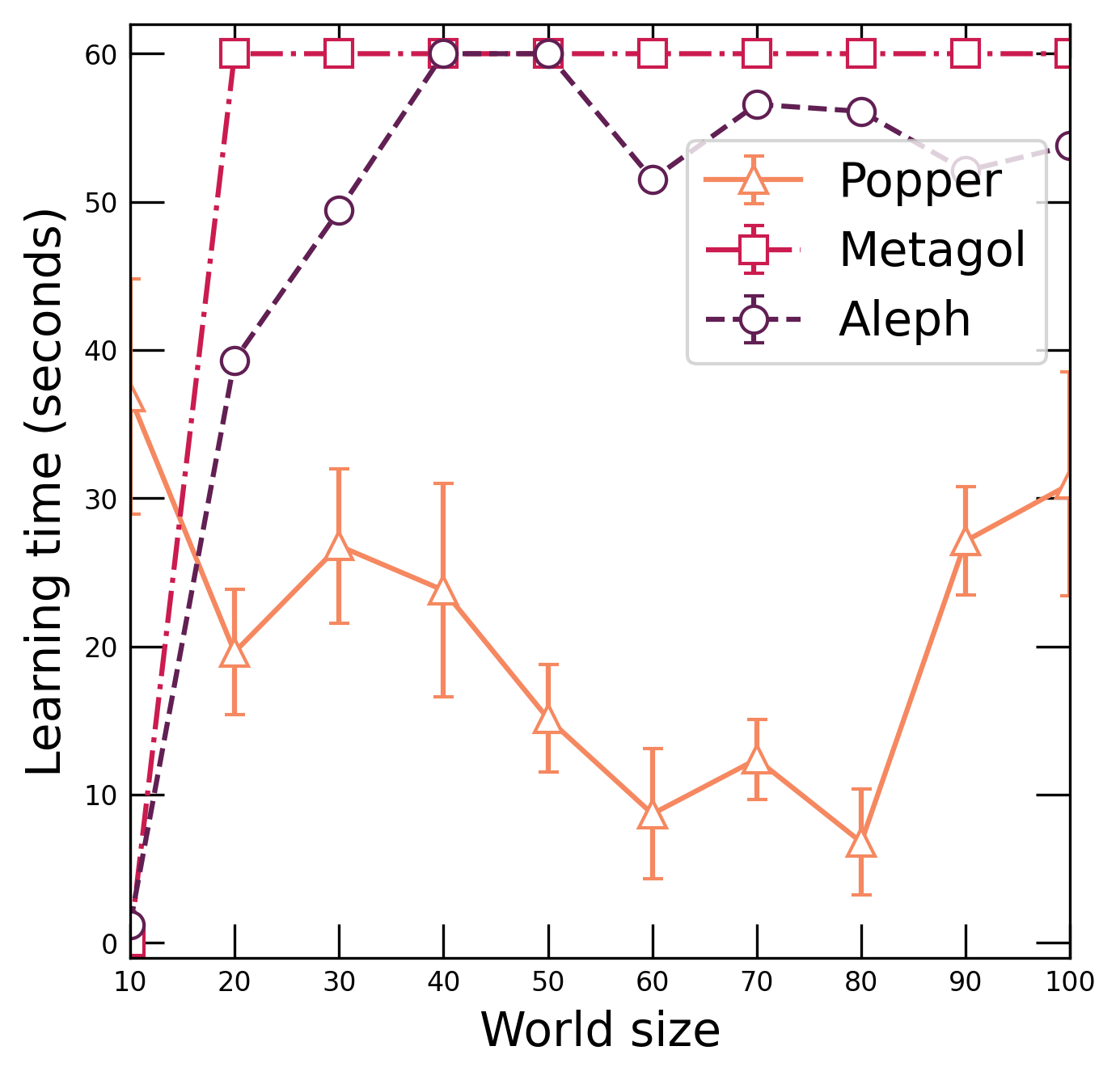}
		\caption{Learning time for $Robot2$ experiment}
		\label{fig:robot2_learning_time}
	\end{figure}
	
	Figure \ref{fig:robot2_accuracy} shows that Popper and Aleph reach 100\% accuracy for all cases, while Metagol struggles due to its inefficient search. Figure \ref{fig:robot2_learning_time} indicates that Popper is the fastest of all the systems. Aleph struggles to learn recursive programs as the grid size becomes larger. Metagol struggles due to its inefficient search, while Popper learns restrictions from failure and never attempts failed clauses again.

    \subsection{Scalability} 
	
    The purpose of this comparison is to check the scalability of each ILP system. To do so, we vary the number of examples in the $member$ problem: check if a number is a member of the given list. We produce $n$ positive and $n$ negative examples, for each $n$ in $[1000, 2000, ..., 10000]$. We repeat each experiment ten times to get the mean and the standard error.
	
	\begin{figure}[h]
		\includegraphics[width=0.5\textwidth,height=0.4\textwidth]{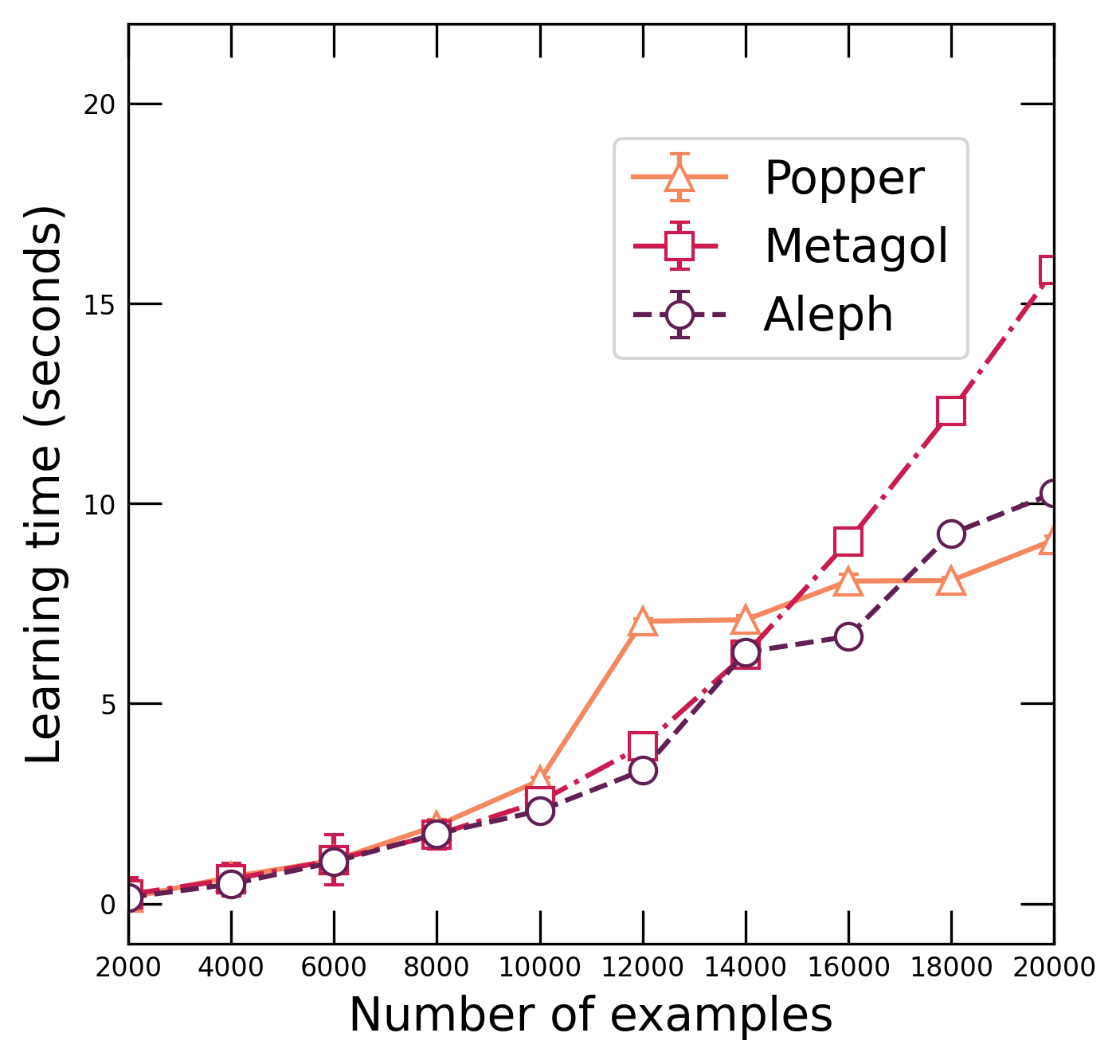}
		\caption{Learning time for $member$ experiment}
		\label{fig:member_learning_time}
	\end{figure}
	
	The results are shown in figure \ref{fig:member_learning_time} when the number of training instances is varied. Note that the accuracies of all three systems are essentially comparable ($100\%$). All systems share similar curves before $10,000$ example size. Aleph and Popper perform better than Metagol after $16,000$ examples.
 
    \subsection{Benchmark ILP problem}
	
    The goal of this comparison is to evaluate how well each system performs on different Benchmark ILP problems. Note that learning recursion and predicate invention are considered two difficult problems for most ILP systems.
	We compare Poper, Metagol, Aleph, $\partial$ILP, and dNL-ILP \cite{payani2019inductive}.
	
	We select four examples from different domains: $grandparent/2$ example from the domain family tree, $even /succ2$ from Arithmetic, $member$ from List, and $undirected$ $edge$ from Graphs. The detailed description of each example is shown in table \ref{table:list_transformation_problems}.
	
    \begin{table*}[]
    \centering
        \begin{tabular}{|c|c|c|c|c|c|}
        \hline
        \textbf{Task}        & \textbf{Description} & \textbf{Domain} & \textbf{$|P_{i}|$} & \textbf{Recursion}        & \textbf{Example solution} \\ \hline
        \textbf{Even}       &    learn even number      & Arithmetic      & 2                  & $\checkmark$ &          	\begin{tabular}[c]{@{}c@{}}$even(X)\leftarrow zero(X)$\\ $even(X)\leftarrow even(Y),pred(Y,X)$\\ $pred(X,Y)\leftarrow succ(X,Z),succ(Z,Y)$ \end{tabular} \\ \hline
        \textbf{Member}      &       Member of a list               & List            & 1                  &                        $\checkmark$   &   
        \begin{tabular}[c]{@{}c@{}}$member(X,Y)\leftarrow value(Y,X)$\\ $member(X,Y)\leftarrow cons(Y,Z), member(X,Z)$  \end{tabular}
        
	                 \\ \hline
         \textbf{Grandparent} &  \begin{tabular}[c]{@{}c@{}}Learn grandparent\\ predicate\end{tabular}   & Family tree    & 2                  &       $\times$                    &          \begin{tabular}[c]{@{}c@{}}$grandparent(X,Y)\leftarrow target(X,Z),target(Z,Y)$\\ $target(X,Z)\leftarrow mother(X,Y)$\\ $target(X,Z)\leftarrow father(X,Y)$ \end{tabular}                 \\ \hline
          \textbf{Undirected edge}     &      \begin{tabular}[c]{@{}c@{}}Learn undirected\\ edge predicate  \end{tabular}           & Graph & 1                  &        $\times$                  &       	
        \begin{tabular}[c]{@{}c@{}}$target(X,Y)\leftarrow edge(X,Y)$\\ $target(X,Y)\leftarrow edge(Y,X)$  \end{tabular}
        
	                  \\ \hline
        \end{tabular}
        \caption{Benchmark ILP problems}
        \label{table:list_transformation_problems}
    \end{table*}
    
    The settings are depicted as follows. For Aleph, the maximum number of nodes to be searched is $50000$. For Metagol, the metarules are the same as those in the $robot$ experiment. The examples should also be provided in increasing sizes to make Metagol find solutions since Metagol is sensitive to the order of examples. There are no restrictions for Popper. 
    
    For each example, we use the same strategy as in the \textit{robot} problem: We generate $100$ positive and $100$ negative examples. The default accuracy is therefore $50\%$. We repeat each experiment ten times to get the mean and the standard error.  
	
    \begin{table*}[]
    \centering
        \begin{tabular}{|c|c|c|c|c|c|}
        \hline
        \textbf{Task}   & \multicolumn{1}{c|}{\textbf{Popper}} & \textbf{Metagol} & \textbf{Aleph} & \textbf{$\partial$ILP} & \textbf{dNL-ILP} \\ \hline
        Even            & $100$                          & $50$       & $50$     & $100$             & $100$      \\ \hline
        Member          & $100$                          & $100$      & $50$    & $100$            & $100$      \\ \hline
        Grandparent     & $100$                          & $100$      & $100$    & $100$            & $100$      \\ \hline
        Undirected edge & $100$                          & $100$      & $100$    & $100$            & $100$      \\ \hline
        \end{tabular}
        \caption{Accuracies for benchmark ILP problems}
        \label{table:list_transformation_problem_predictive_accuracy}
    \end{table*}

    Table \ref{table:list_transformation_problem_predictive_accuracy} indicates that
    Popper, $\partial$ILP, and dNL-ILP perform perfectly on all the tasks in terms of accuracy. Metagol reaches $100\%$ predictive accuracy except for the $even$ problem. Aleph struggles to learn solutions for $even$ and $member$ problems. Table \ref{table:list_transformation_learning_time} shows that Aleph is the fastest and $\partial$ILP is the slowest within all the systems. Note that the main difference between $\partial$ILP and dNL-ILP is that $\partial$ILP allows for clauses of at most two atoms and only two rules per each predicate to reduce the size of the search space, while in dNL-ILP, the membership weights of the conjunction can be directly interpreted as the flags in the satisfiability interpretation \cite{payani2019inductive}.

    \begin{table*}[]
        \centering
        \begin{tabular}{|c|c|c|c|c|c|}
        \hline
        \textbf{Task}   & \multicolumn{1}{c|}{\textbf{Popper}} & \textbf{Metagol} & \textbf{Aleph} & \textbf{$\partial$ILP} & \textbf{dNL-ILP} \\ \hline
        Even            & $1.2 \pm  0.08$                      & $1.5 \pm 0.12$    & $0.03 \pm 0$   & $1332.56 \pm 33.90$    & $10.43 \pm 1.02$ \\ \hline
        Member          & $3 \pm  0.17$                        & $0.02 \pm 0$     & $0.02 \pm 0$   & $849.25 \pm 63.35$     & $8.27 \pm 0.66$ \\ \hline
        Grandparent     & $3.5 \pm  0.21$                        & $0.02 \pm 0$     & $0.03 \pm 0$   & $1097.50 \pm 53.60$    & $17.12 \pm 0.93$ \\ \hline
        Undirected edge & $4.2 \pm  0.38$                       & $3.5 \pm 0.33$    & $0.02 \pm 0$   & $469.31\pm 26.04$      & $12.28 \pm 1.16$ \\ \hline
        \end{tabular}
        \caption{Learning times for benchmark ILP problems}
        \label{table:list_transformation_learning_time}
    \end{table*}
	
    \subsection{Feature comparison}
	
	We compare six ILP systems in terms of the inference rules they employ. In the comparison, We compare important features, including 1) handling recursive rules: a logic program can handle recursive rules if the same predicate appears in the head and body of a rule, 2) robustness: the ability of a system to cope with errors during execution and cope with erroneous input \cite{fernandez2005model}, 3) predicate invention: a way of finding new theoretical terms, or abstract new concepts that are not directly observable in the measured data \cite{muggleton2012ilp}, and (4) scalability: it describes the capability of a system to cope and perform well under an increased or expanding workload or scope. Although this list can be extended with other features, we aim to include features that we believe are important for different ILP systems. The comparison of the systems is depicted in Table \ref{table:feature_comparison_for_ilp_system}. 

    \begin{table*}[]
    \centering
        \begin{tabular}{|c|c|c|c|c|c|c|}
        \hline
        \textbf{Task}      & \textbf{Aleph}                                                & \textbf{Metagol}                                               & \textbf{Popper}                                                & \textbf{ProbLog}                                         & \textbf{DILP}                                                         & \textbf{dNL-ILP}                                                      \\ \hline
        \textbf{Inference Rule} & \begin{tabular}[c]{@{}c@{}}Inverse \\ Entailment\end{tabular} & \begin{tabular}[c]{@{}c@{}}High-order\\ Abduction\end{tabular} & \begin{tabular}[c]{@{}c@{}}High-order\\ Abduction\end{tabular} & \begin{tabular}[c]{@{}c@{}}SLD\\ Resolution\end{tabular} & \begin{tabular}[c]{@{}c@{}}High-order\\ Abduction\\ + NN\end{tabular} & \begin{tabular}[c]{@{}c@{}}High-order\\ Abduction\\ + NN\end{tabular} \\ \hline
        \textbf{Recursion}      & Limited                                                       & $\checkmark$                                                   & $\checkmark$                                                   & Limited                                                  & $\checkmark$                                                          & $\checkmark$                                                          \\ \hline
        \textbf{Robustness}      & $\times$                                                      & $\times$                                                       & $\times$                                                       & $\checkmark$                                             & $\checkmark$                                                          & $\checkmark$                                                          \\ \hline
        \textbf{Predicate Invention} & $\times$                                                      & $\checkmark$                                                   & $\checkmark$                                                   & $\times$                                                 & $\checkmark$                                                          & $\checkmark$                                                          \\ \hline
        \textbf{Scalability}     & Limited                                                      & Limited                                                       & Good                                                       & Good                                             & Bad                                                              & Good                                                              \\ \hline
        \end{tabular}
		\caption{Feature comparison for ILP systems}
		\label{table:feature_comparison_for_ilp_system}        
    \end{table*}
	
	Recursive clauses are part of the hypothesis in some learning examples. In a recursive rule, one atom at least shows in both the head and the body. Although the ability to generate recursive rules is important to a system, it is often easier for an ILP system to generalize from small numbers of examples with recursion \cite{cropper2015meta}. All models in Table \ref{table:feature_comparison_for_ilp_system} could generate recursive rules, but traditional ILP systems struggle to learn recursive programs, especially from small numbers of training examples \cite{cropper2020turning}. For the robustness of each system, all the traditional ILP systems, as well as Metagol and Popper, can not be robust to noise in and mislabelled inputs. As PILP combines uncertainty with logic, while dNL-ILP and $\partial$ILP connect ILP with neural networks, they can deal with noise and ambiguity.
	
	Predicate invention in ILP means the automatic introduction of new and hopefully useful predicates during learning from examples. Such new predicates can then be used as part of background knowledge in finding a definition of the target predicate \cite{muggleton2012ilp}. Traditional ILP systems have nonsupport for predicate invention. Since Metagol and Popper use high-order metarules to define the hypothesis space, the new predicate will be invented when induce definitions are necessary. $\partial$ILP and dNL-ILP also use rules as guidance to support predicate invention.
	
	Existing ILP systems and Metagol can not be applied effectively for data sets with 10000 data points for each system's scalability. Approximative Generalization can compress several data points into one example to tackle large datasets in PILP \cite{watanabe2009can}. $\partial$ILP is restricted to small datasets because of its memory requirements. The scalability of $\partial$ILP is not satisfied due to the predicates constraints (only two predicates in each clause are allowed in $\partial$ILP to avoid the memory-consuming problem). It should be good if the memory-consuming problem can be solved since gradient descent used in $\partial$ILP performs well under an increased workload.
	
	\subsection{Summarize}
    We test several systems in this section by adjusting: 1) optimal solution size, 2) optimal domain size, 3) example size, and 4) benchmark ILP problems. We broadly summarize the learning times and predictive accuracies of all models in the experiments: Although Aleph is faster than all the other systems in benchmark ILP problems, it only learns accurate solutions for $Grandparent$ and $Undirected$ $edge$. Popper can be roughly considered an upgrade of Metagol: by learning from failure, Popper finds accurate solutions with small time-consuming for most of the experiments. The performance of Popper highly relies on initial restrictions, while The performance of Metagol heavily depends on metarules. $\partial$ILP and dNL-ILP only handle Benchmark ILP problems, although they produce reliable results. 

    \begin{table*}[]
        \centering
        \begin{tabular}{|c|c|c|c|c|c|c|}
        \hline
        \multicolumn{2}{|c|}{\textbf{Task}}                                      & \textbf{Aleph} & \textbf{Metagol} & \textbf{Popper} & \textbf{$\partial$ILP} & \textbf{dNL-ILP} \\ \hline
        \multicolumn{2}{|c|}{\textbf{Robot}}                                   & slow           & fast             & slow            & x             & x                \\ \hline
        \multicolumn{2}{|c|}{\textbf{Robot2}}                                    & medium         & slow             & fast            & x             & x                \\ \hline
        \multirow{4}{*}{\textbf{Benchmark ILP}} & \textbf{Even}            & fast           & medium           & medium          & slow          & medium           \\ \cline{2-7} 
                                                      & \textbf{Member}          & fast           & fast             & medium          & slow          & medium           \\ \cline{2-7} 
                                                      & \textbf{Grandparent}     & fast           & fast             & medium          & slow          & medium           \\ \cline{2-7} 
                                                      & \textbf{Undirected edge} & fast           & medium           & medium          & slow          & medium           \\ \hline
        \end{tabular}
        \caption{Learning times for experiments}
        \label{table:learning_time_for_experiments}
    \end{table*}
    
    \section{Challenges} 
	\label{sec:challenges}

    In this section, We share challenges facing the community based on the recent developments surveyed in this paper:
    
    \paragraph{Data efficiency} Symbolic methods deal with small amounts of data efficiently, while the leading approaches in neural models need large amounts of data to achieve outstanding performance. Meanwhile, symbolic methods have difficulty dealing with large datasets, but neural models address them easily. For SRL, the structure learning, i.e., in MLNs, is not scalable and inefficient for large amounts of data as well. This has triggered the development of NeSy and SRL. To learn the benefits of each model and overcome their complimentary flaws is a potential research direction.    
    
    \paragraph{Generalization} Generalization refers to the capacity of models to adapt to new, previously unseen data derived from the same distribution as the one used to create the model. Understanding generalization is one of the fundamental unsolved problems for all machine learning models. We found that several symbolic-based models suffer from weak generalization. For instance, Metagol sometimes can not generate valid inference rules when changing the number range from ten to twenty for positive and negative examples in the $even$ dataset. Although the search method during training can be fully guided and understood, enough attention should be paid to the failure of generalization that may be caused only by a tiny change in the input.
    
    \paragraph{Application} 
    From an application point of view, the usage of traditional symbolic-based applications is very limited, and applications in SRL and NeSy are rarely developed. As the integration of symbolic with probability and neural networks, hybrid methods can be adapted to new fields such as computer vision, natural language processing, etc. As explainability and logic play crucial roles in many fields, methods in SRL and NeSy will be challenging and have great potential for the field.

    \paragraph{Explainability} Symbolic AI holds certain merits in explainability due to its symbolic nature. Although AI users will increasingly require explanations, the current research in symbolic AI and the related integrations need to pay more attention to the target users. Recent work \cite{ribera2019can} shows a new explainability pipeline, which can be applied to symbolic AI. Another work \cite{ai2020beneficial} investigates the explanatory effects of a machine-learned theory in the context of simple two-person games and proposes a framework for identifying the harmfulness of machine explanations based on the Cognitive Science literature. However, more work is required for explanations in machine-learned theories, and the creation of different user-centered explainability solutions is required.
	
	\section{Conclusion}
	\label{sec:conclusion}
 
    The increasing interests in symbolic AI and symbolic-based integration have been shown by various communities. This survey presents an overview of the symbolic family of algorithms, especially ILP. Traditional ILP holds certain merits on machine learning algorithms' explainability due to its symbolic nature and variants of ILP, e.g., PILP, MIL, and $\partial$ILP maintain interpretations and add powerful new features. The pros and cons of related research areas, such as statistical relational learning and Neural-Symbolic AI, are also investigated. Combining user-centered explainable AI with ILP is also reviewed, as we consider user-centered design the next frontier of AI. We believe that the character and development of integrated symbolic AI will have a significant impact on AI in the future.

%% file: biography.tex

\vspace{-20 mm}

\begin{IEEEbiography}[{\includegraphics[width=1in,height=1.25in,clip,keepaspectratio]{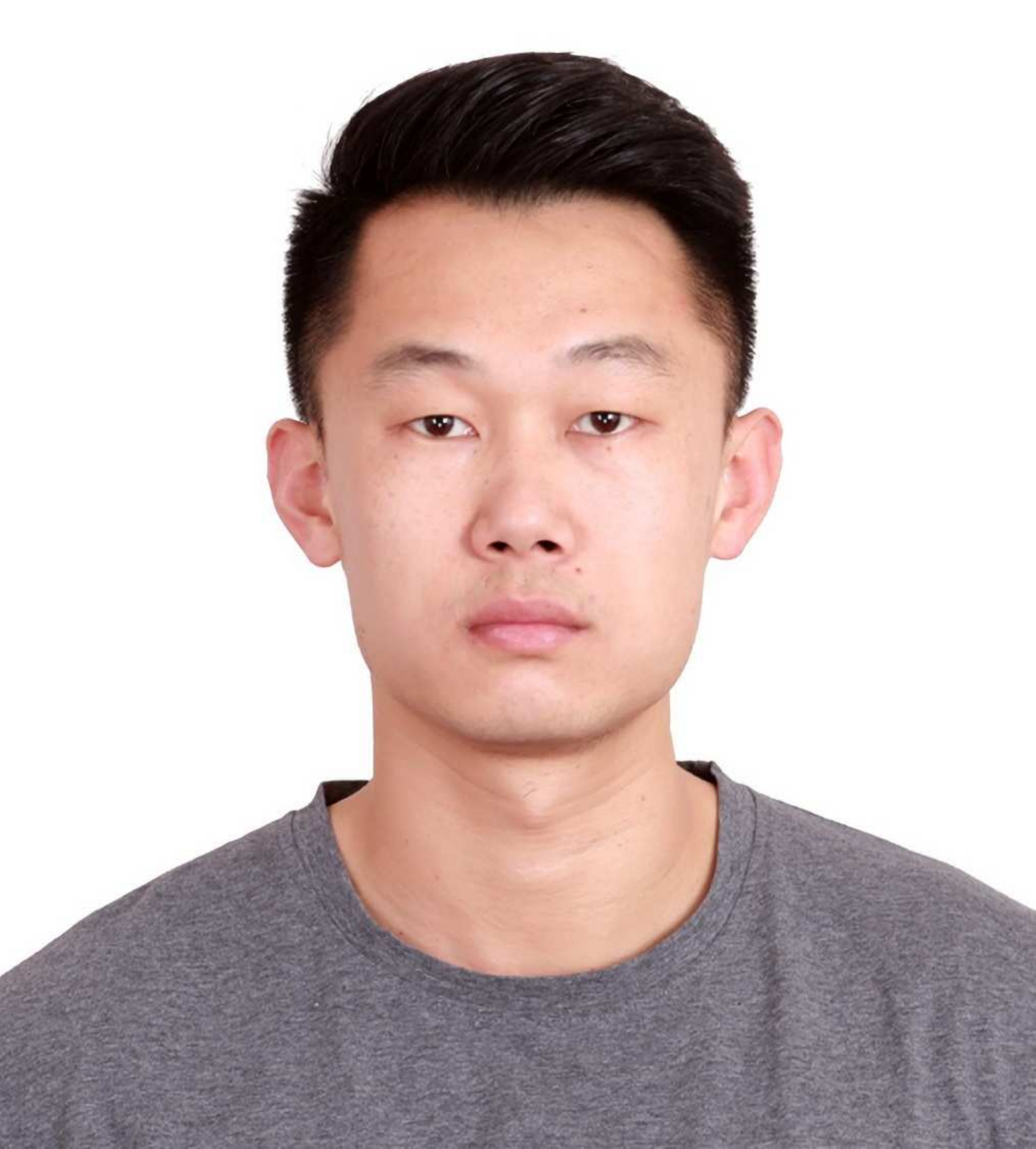}}]{Zheng Zhang}
received his M.S.degree in Electrical Engineering from New Jersey Institute of Technology, and M.S.degree in Computer Science from Auburn University, USA, in 2014 and 2018, respectively. He is currently a Ph.D. student in the Department of
Computer Science and Software Engineering, Auburn University, USA. His research interests include Neuro-symbolic AI, reinforcement learning, computer vision, and artificial intelligence.
\end{IEEEbiography}

\vspace{-20 mm}

\begin{IEEEbiography}[{\includegraphics[width=1in,height=1.25in,clip,keepaspectratio]{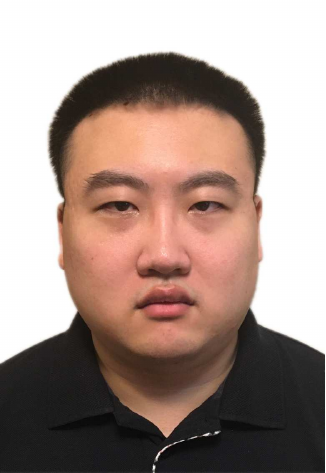}}]{Liangliang Xu}
received his M.S.degree in Industiral Engineering and Computer Science from Auburn University, USA, in 2018, respectively. He is currently a Ph.D. student in the Department of
Computer Science and Software Engineering, Auburn University, USA. 
\end{IEEEbiography}

\vspace{-20 mm}

\begin{IEEEbiography}[{\includegraphics[width=1in,height=1.25in,clip,keepaspectratio]{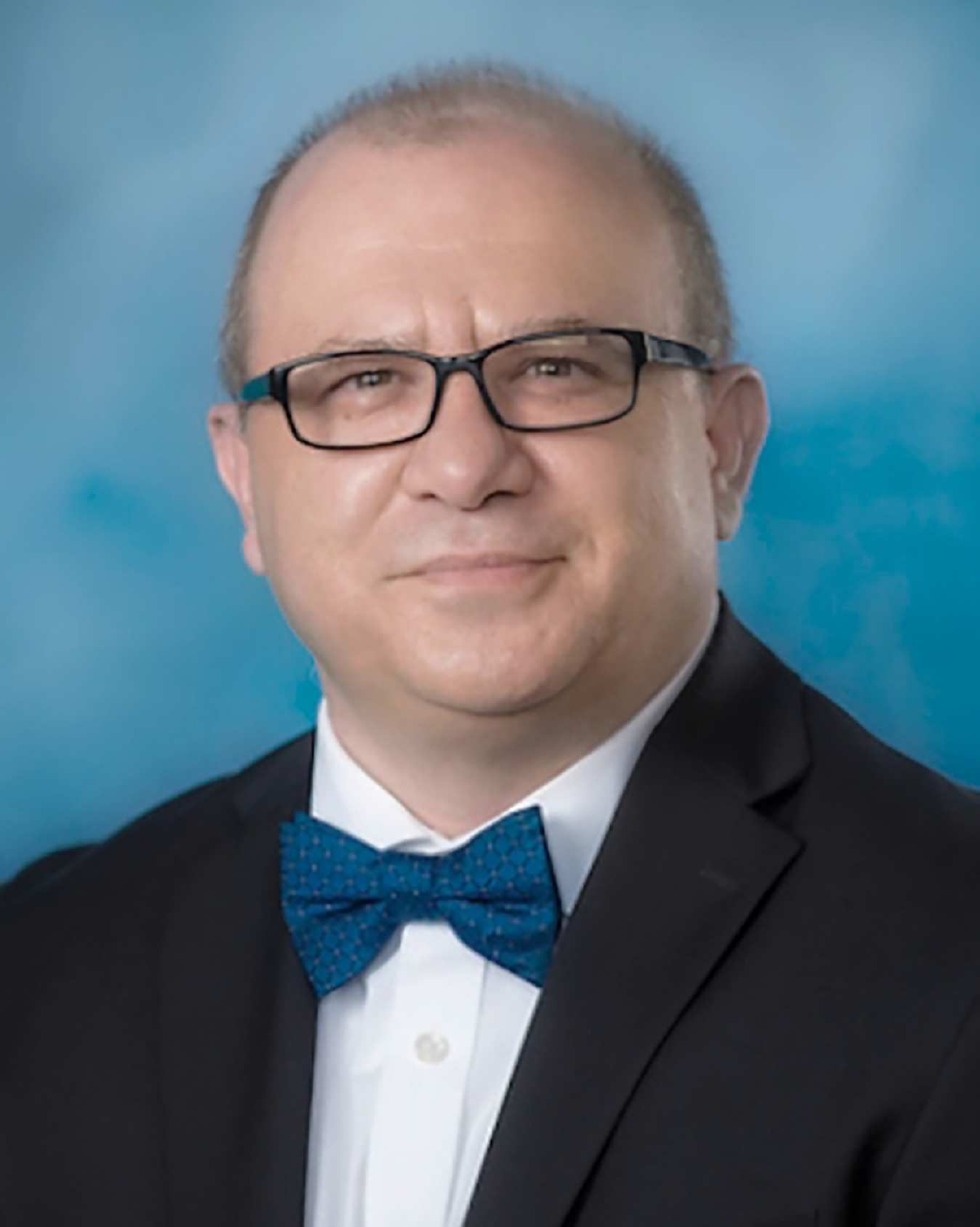}}]{Levent Yilmaz}
is  the Alumni Distinguished Professor of Computer Science and Software Engineering at Auburn University with a courtesy appointment in Industrial and Systems Engineering.
He holds M.S. and Ph.D. degrees in Computer Science from Virginia Tech.
His research interests are in theory and methodology of modeling and simulation, agent-directed simulation, cognitive systems, and model-driven science and engineering for complex adaptive systems.
He is the former Editor-in-Chief of Simulation: Transactions of the Society for Modeling and Simulation International and the founding organizer and general chair of the Agent-Directed Simulation Conference series.
\end{IEEEbiography}

\vspace{-20 mm}

\begin{IEEEbiography}[{\includegraphics[width=1in,height=1.25in,clip,keepaspectratio]{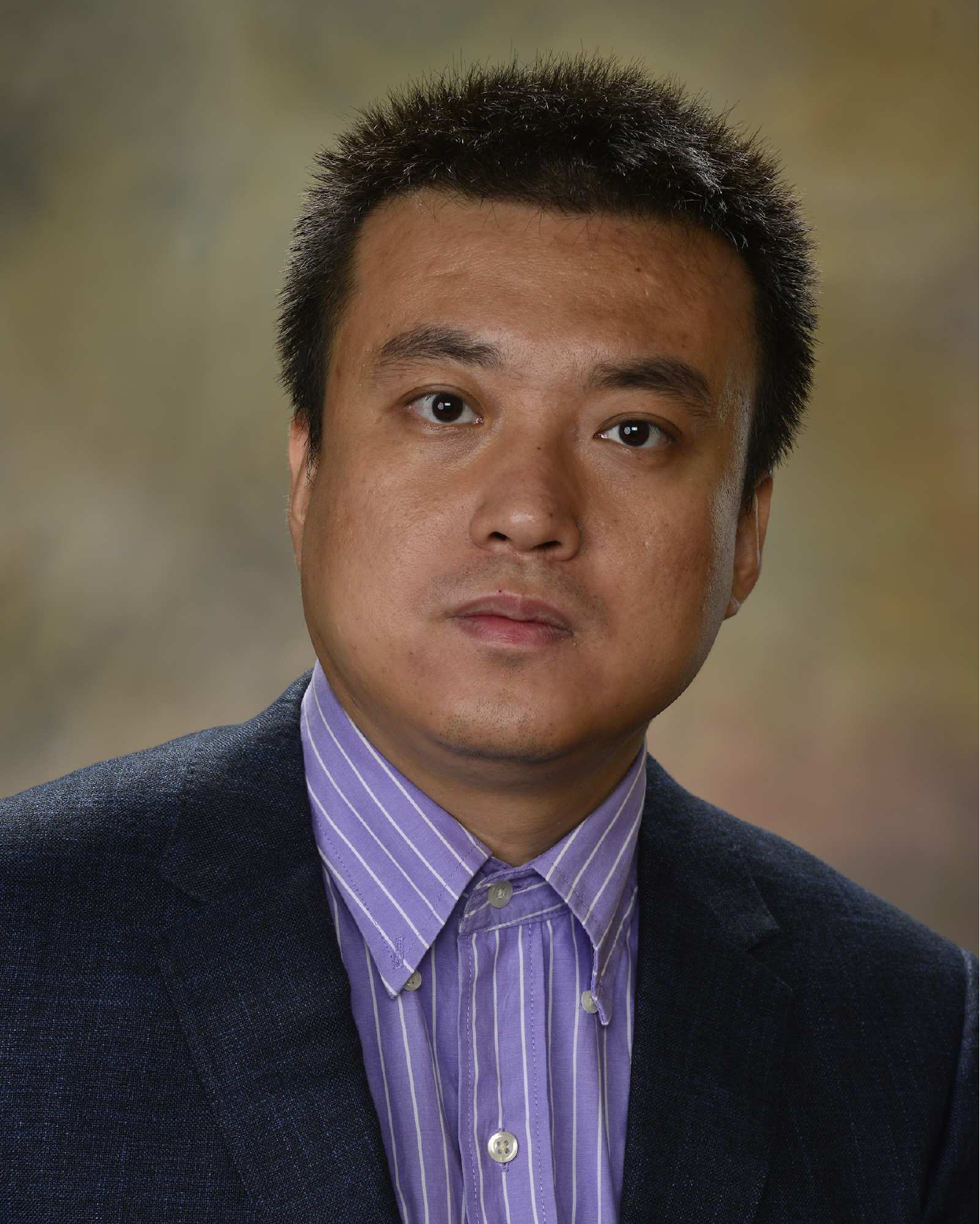}}]{Bo Liu} is a Tenure-Track Assistant Professor with the Department of Computer Science and Software Engineering, Auburn University, Auburn, AL, USA. He received the Ph.D. degree from the University of Massachusetts Amherst, Amherst, MA, USA, in 2015. He has over 30 publications on several notable venues. He is the recipient of the UAI'2015 Facebook best student paper award and the Amazon research award in 2018.  He is an Associate Editor of IEEE Transactions on Neural Networks and Learning Systems (IEEE-TNN), a senior member of IEEE, and a member of AAAI, ACM, and INFORMS.
\end{IEEEbiography}